\RequirePackage{fix-cm}
\documentclass[twocolumn]{svjour3}          % twocolumn
\smartqed  % flush right qed marks, e.g. at end of proof
\usepackage{graphicx}
\usepackage{amsmath}
\usepackage{amssymb} % define this before the line numbering.
\usepackage{color}
\usepackage{times}
\usepackage{epsfig}
\usepackage{subfig}
\usepackage{booktabs}
\usepackage{enumitem}
\usepackage{amsmath}
\usepackage{multirow}
\newcommand{\change}[1]{\textcolor{black}{#1}}
\usepackage{tikz}
\def\checkmark{\tikz\fill[scale=0.4](0,.35) -- (.25,0) -- (1,.7) -- (.25,.15) -- cycle;} 

\begin{document}
\sloppy

\title{Adversarial Framework for Unsupervised Learning of Motion Dynamics in Videos%\thanks{Grants or other notes
}

\author{C. Spampinato \and S. Palazzo \and P. D'Oro \and D. Giordano \and M. Shah%etc.
}


\institute{C. Spampinato, S. Palazzo, P. D'Oro, D. Giordano \at
              PeRCeiVe Lab - University of Catania - Italy \\
              Tel.: +39-0957387902\\
              \email{\{cspampin, palazzosim, dgiordan@dieei.unict.it\}}           %  \\
           \and
           M. Shah and C. Spampinato \at
           Center for Research in Computer Vision - University of Central Florida\\
           Tel.: +1 (407) 823-1119\\
           \email{shah@crcv.ucf.edu}
}
%\date{}
\maketitle
%%%%%%%%% ABSTRACT
\begin{abstract}
Human behavior understanding in videos is a complex, still unsolved problem and requires to accurately model motion at both the local (pixel-wise dense prediction) and global (aggregation of motion cues) levels. Current approaches based on supervised learning require large amounts of annotated data, whose scarce availability is one of the main limiting factors to the development of general solutions. Unsupervised learning can instead leverage the vast amount of videos available on the web and it is a promising solution for overcoming the existing limitations.
In this paper, we propose an adversarial GAN-based framework that learns video representations and dynamics through a self-supervision mechanism in order to perform dense and global prediction in videos. Our approach synthesizes videos by 1) factorizing the process into the generation of static visual content and motion, 2) learning a suitable representation of a motion latent space in order to enforce spatio-temporal coherency of object trajectories, and 3) incorporating motion estimation and pixel-wise dense prediction into the training procedure. Self-supervision is enforced by using motion masks produced by the generator, as a co-product of its generation process, to supervise the discriminator network in performing dense prediction. Performance evaluation, carried out on standard benchmarks, shows that our approach is able to learn, in an unsupervised way, both local and global video dynamics. The learned representations, then, support the training of video object segmentation methods with sensibly less (about 50\%) annotations, giving performance comparable to the state of the art. 
Furthermore, the proposed method achieves promising performance in generating realistic videos, outperforming state-of-the-art approaches especially on motion-related metrics.
\end{abstract}

%%%%%%%%% BODY TEXT
\section{Introduction}
Learning motion dynamics plays an important role in video understanding, which fosters many applications, such as object tracking, video object segmentation, event detection and human behavior understanding. The latter is a particularly complex task, due to the variability of possible scenarios, conditions, actions/behaviors of interest, appearance of agents and to a generic ambiguity in how behaviors should be defined, categorized and represented. Behavior understanding is also a key component in the direction toward visual intelligence: the identification of what happens in a given environment necessarily requires the capability to decode actions and intentions from visual evidence.

If behavior understanding in computer vision is a fundamental component  for scene understanding, pixel-wise
dense prediction for video object segmentation is one of the founding stones for the whole process, as it isolates relevant regions in a scene from unnecessary background elements, thus serving both as a way to focus analysis on a subset of the input data, and to compute a preliminary representation suitable for further processing. Unfortunately, although video object segmentation has been studied for decades, it is far from solved, as current approaches are not yet able to generalize to the variety of unforeseeable conditions that are found in real-world applications. Additionally, learning long-term spatio-temporal features directly for dense prediction greatly depends on the availability of large annotated video object segmentation benchmarks (e.g., the popular DAVIS 2017 benchmark dataset  contains only 150 short video clips, barely enough for training end-to-end deep models from scratch).
The alternative approach is to avoid resorting to manual annotations by explicitly defining a ``cost'' or ``energy'' function based on \textit{a priori} considerations on the motion patterns that characterize objects of interest in a video~\cite{papazoglou2013fast,7299114}. However, these approaches do not seem to be on par with deep learning methods.

Recently, generative adversarial networks~(GANs)~\cite{NIPS2014_5423} have become a successful trend in computer vision and machine learning, thanks to their results in advancing the state of the art on image generation to unprecedented levels of accuracy and realism~\cite{NIPS2015_5773,RadfordMC15,Zhang_2017_ICCV,pmlr-v70-arjovsky17a,Mao_2017_ICCV,NIPS2017_6797,Huang_2017_CVPR}. The key idea of GANs is to have two models, a \emph{generator} and a \emph{discriminator}, compete with each other in a scenario where the discriminator learns to distinguish between real and fake samples, while the generator learns to produce more and more realistic images. As the models improve in parallel, they learn hierarchies of general feature representations that can be used for multiple tasks, e.g., image classification~\cite{RadfordMC15} and semantic segmentation~\cite{Souly_2017_ICCV}. These characteristics have demonstrated GANs' usefulness in training or supporting the training of models from unlabeled data~\cite{NIPS2016_6125}, rising as one of the most promising paradigms for unsupervised learning\footnote{Note the distinction between \emph{unsupervised learning} (the class of approaches that train machine learning models without annotations) and what is known in the video segmentation literature as \emph{unsupervised segmentation} (the class of methods that perform segmentation in inference without additional input on object location other than the video frames).}. 

Given the success of GANs with images, the natural direction of research has been to attempt and extend their applicability to videos, both as a generative approach and as a way to disentangle video dynamics by learning features that leverage the vast amount of unlabeled data available on the web. For example, in a classification scenario such as \emph{action recognition}, a simple way to employ unlabeled videos is to add an additional class for fake videos (i.e., produced by the generator) and have the discriminator both predict the realism of the input video and identify its class~\cite{odena2016semi,NIPS2016_6125}. However, naively extending image generation methods to videos by adding the temporal dimension to convolutional layers may be too simplistic, as it jointly attempts to handle both the spatial component of the video, which describes object and background appearance, and the temporal one, representing object motion and consistency across frames. Building on these considerations, recent generative efforts~\cite{NIPS2016_6194,Saito_2017_ICCV} have attempted to factor the latent representation of each video frame into two components that model a time-independent background of the scene and the time-varying foreground elements. We argue that the main limitation of these methods is that both factors are learned by mapping a single point of a latent space (sampled as random noise) to a whole video. This, indeed, over-complicates the generation task as two videos depicting the same scene with different object trajectories or the same trajectory on different scenes are represented as different points in the latent space, although they share a common factor (in the first case the background, in the second one object motion).

\begin{figure*}[h]
	\centering
	\includegraphics[width=1\textwidth]{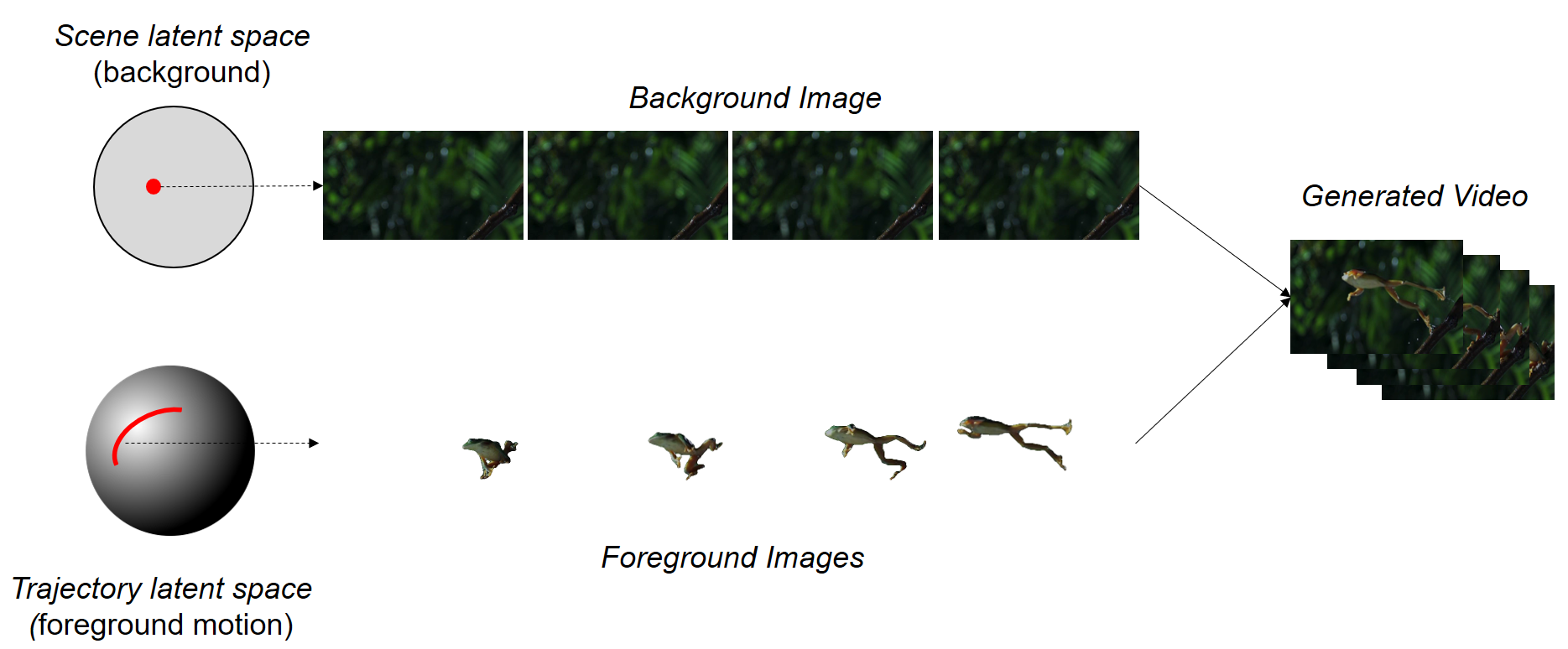}
	\caption{\textbf{Our adversarial video generation model}: we employ a scene latent space to generate background and a foreground latent space to generate object appearance and motion.}
	\label{fig:method}
\end{figure*}

In this paper, we tackle both the problem of unsupervised learning for video object segmentation and that of video generation with disentangled background and foreground dynamics, combining both of them into an adversarial framework that guides the discriminator in performing video object segmentation through a self-supervision mechanism, by using ground-truth masks internally synthesized by the generator. 
%Our framework mainly relies on a GAN-based video generation deep model, which, internally, generates segmentation masks that are employed for supervising pixel-wise dense prediction.
In particular, our video generation approach employs two latent spaces (as shown in Fig.~\ref{fig:method}) to improve the video generation process: 1) a traditional random latent space to model the static visual content of the scene (background), and 2) a trajectory latent space suitable designed to ensure spatio-temporal consistency of generated foreground content. In particular, object motion dynamics are modeled as point trajectories in the second latent space, with each point representing the foreground content in a scene and each latent trajectory ensuring regularity and realism of the generated motion across frames. %Variations in the scene latent space are related to different scenes, while variations in the trajectories of the foreground latent space are related to different object motion. 
On top of the traditional adversarial framework, we extend the discriminator architecture in order to perform adversarial dense pixel-wise prediction in videos. In particular, besides the adversarial loss driving the generator/discriminator game, we add loss terms related to the discriminator's estimation of optical flow (supervised by the output of a state-of-the-art algorithm) and segmentation masks (supervised by the foreground masks computed by the generator) from the generated videos. The three losses encourage the generator to produce realistic videos, while improving representation learning in the discriminator and unlocking the possibility to perform dense video predictions with no manual annotations.
Experimentally, we verify that our video generation approach is able to effectively synthesize videos, outperforming existing solutions, especially in motion coherency metrics, thus suggesting that it indeed learns, in an unsupervised way, motion features. 
We further demonstrate that the features learned by the model's discriminator can be used for effective unsupervised video object segmentation in different domains and allow for reducing significantly (about 50\% less) the number of annotated frames required to achieve the same performance as through traditional supervision. Additionally, we find that the features learned through unsupervised learning encode general appearance and motion cues and can be also employed for global prediction tasks such as video action recognition.

To summarize, the main contributions of this paper are:
\begin{itemize}
	\item We introduce a GAN-based video generation framework able to explicitly model object motion through learning a latent trajectory representation space that enforces spatio-temporal regularity of object motion in the generated videos as well estimating motion cues from the generated content;
	\item We demonstrate that our framework provides a useful means for video object segmentation --- known as being an annotation-hungry task --- by both employing a trained generator to create synthetic foreground masks and directly integrating dense prediction into the adversarial training procedure;
	\item We verify that our approach is able to successfully learn video features that can be used in a variety of computer vision tasks.
\end{itemize}

\section{Related Work}
This paper mainly tackles the problem of unsupervised learning of motion dynamics for pixel-wise dense prediction in a video object segmentation scenario through an adversarial video generation framework. Thus, we first review the recent literature on video object segmentation methods and then focus on video generation approaches. 

Video object segmentation is the task of predicting each video pixel either as foreground or background and consequently to segment objects preserving their boundaries. Recent video object segmentation methods can be classified as \emph{unsupervised}, i.e., methods that perform segmentation in inference without additional input on object location other than the video frames, \emph{semi-supervised}, i.e., methods that, instead, employ annotations in the first frame for inference, and \emph{supervised}, i.e. methods that require annotations for every frame or user interaction.
In this paper we propose a framework that enables to learn motion features through \emph{unsupervised learning}, i.e., without using annotations at training time. The learned representations are then employed to train a method for unsupervised video object segmentation; thus, in this section, we will focus on this last class of approaches.

Many of these methods formulate the problem as a spatio-temporal tube (representing motion trajectories) classification task \cite{papazoglou2013fast,Brox:2010,6247883,6126471,keuper2015,7298961}. The core idea is to track points or regions over consecutive frames in order to create coherent moving objects, by learning motion/appearance cues. Brox and Malik \cite{Brox:2010} propose a pairwise metric on trajectories across frames for clustering trajectories. Analogously, Fragkiadaki et al. \cite{6247883}  analyze embedding density variations between spatially contiguous trajectories to perform segmentation. Papazoglou and Ferrari \cite{papazoglou2013fast}, instead, model motion, appearance and spatial feature density forward and backward in time across frames to extract moving objects. Keuper et al. \cite{keuper2015} also track points in consecutive frames and employ a multicut-based approach for trajectory clustering. Wang et al. \cite{7298961} define spatio-temporal saliency, computed as geodesic distance of spatial edges and temporal motion boundaries, and employ such saliency as a prior for segmentation. 

Recently, CNN-based methods for unsupervised video object segmentation have been proposed \cite{8099855,JainXG17,Cae17,Tokmakov17}, based on the concept of making the network implicitly learn motion/appearance cues for object trajectory modeling. Most of CNN-based methods use pre-trained segmentation models \cite{8099855,JainXG17,Cae17} or optical-flow models \cite{Tokmakov17} to perform either semi-supervised or unsupervised segmentation. 
Such methods are, however, profoundly different from the more challenging case of learning to segment moving objects without utilizing, or reducing significantly the need of, labeled data at training time. Thus, although our approach shares the strategy of these methods, i.e., making a CNN learn motion and appearance cues for segmentation, it significantly differs from them in that we learn motion cues in a completely unsupervised way during training. 

To carry out unsupervised learning, we leverage the recent success of generative adversarial networks (GANs) \cite{NIPS2014_5423} for high-quality image \cite{NIPS2015_5773,RadfordMC15,Zhang_2017_ICCV,NIPS2015_5773,pmlr-v70-arjovsky17a,Mao_2017_ICCV,NIPS2017_6797}, video synthesis \cite{NIPS2016_6194,Saito_2017_ICCV,Tulyakov_2018_CVPR,ohnishi2018hierarchical} and prediction \cite{Jang2018VideoPW,Vondrick_2017_CVPR}, by devising a video generation model that performs, at the same time, pixel-wise dense prediction for video object segmentation. Existing adversarial deep image generation methods have attempted to increase the level of realism \cite{NIPS2015_5773,RadfordMC15,Zhang_2017_ICCV,NIPS2015_5773} of the generated images, while providing solutions to cope with training stabilization issues \cite{pmlr-v70-arjovsky17a,Mao_2017_ICCV,NIPS2017_6797}.
Most of these methods use a single input latent space to describe all possible visual appearance variations, with some exceptions, such as Stacked GAN \cite{Huang_2017_CVPR} that, instead, employs multiple latent spaces for different image scales. Works reported in \cite{NIPS2016_6194,Saito_2017_ICCV,Tulyakov_2018_CVPR} extend the GAN image generation framework to the video generation problem. In particular, the work in \cite{NIPS2016_6194} replaces the GAN image generator with a spatio-temporal CNN-based network able to generate, from an input latent space, background and foreground separately, which are then merged into a video clip for the subsequent discriminator. Similarly, the work in \cite{Saito_2017_ICCV} shares the same philosophy of \cite{NIPS2016_6194} with the difference that the input latent space is mapped into a set of subspaces, where each one is used for the generation of a single video frame. However, simply extending the traditional GAN framework to videos fails, because of the time-varying pixel relations that are found in videos due to structured object motion. Although the above generation methods exploit video factorization into a stationary and a temporally varying part, deriving the two components from a single input latent space greatly complicates the task, and needs more training data given that each point in the input latent space corresponds to a complete scene with specific object motion. Furthermore, object motion is only loosely linked to the scene, e.g., the motion of a person walking on two different environments is more or less independent from the specific environment. Thus, the assumption that the two video components are highly inter-correlated (so as to derive both from a single point in the latent space) is too strong, and recently \cite{Tulyakov_2018_CVPR} performed video generation by disentangling the two video components into different latent spaces. 
While our approach is in the same spirit of \cite{Tulyakov_2018_CVPR}, there are several crucial differences both in the problem formulation and in the motivation. 
In terms of motivation, our approach performs video generation with the main goal to learn motion dynamics in an unsupervised manner, and it is designed in order to implement a self-supervision mechanism for guiding the pixel-wise dense prediction process. In terms of problem formulation, we use two latent spaces as in \cite{Tulyakov_2018_CVPR}, but our foreground latent space is a multidimensional space from which we sample trajectories, and not a sequence of uncorrelated and isolated points, as done in \cite{Tulyakov_2018_CVPR}. These trajectories are fed to a recurrent layer that learns suitable temporal embeddings.
Thus, we learn a latent representation for motion trajectories, which, as shown in the results, leads to a better spatio-temporal consistency than \cite{Tulyakov_2018_CVPR}.
\change{Along the line of using two separate latent spaces for background and foreground modeling, \cite{OhnishiYUH18} propose a video generation approach that first generates optical flow and then converts it to foreground content for integration with a background model. Our approach differs from this work in the following ways:
\begin{itemize}
    \item Although \cite{OhnishiYUH18} employs two separate latent spaces for motion and content, single samples are drawn from the two for generating a video; instead, we learn a motion latent space, which more naturally maps to the spatio-temporal nature of motion, and encodes it as input for the generation process in the foreground stream.
    \item It employs optical flow provided by a state-of-the-art algorithm as a condition, as done in standard conditional GANs, in the hierarchical structure of the generator. We, instead, estimate optical flow through the discriminator and use it to supervise, in the form of ``self-supervision'', video generation in order to directly encourage a better understanding of motion dynamics.
    \item Lastly, this work  does not address the video object segmentation problem, which remains the main objective of our work.
\end{itemize}}

\change{Video GANs have been also adopted for the video prediction task \cite{Vondrick017,VillegasYHLL17,pmlr-v80-jang18a}, i.e., for conditioning the generation process of future frames given a set of input frames. While this is a different task than what we here propose, i.e., video generation for supporting spatio-temporal dense segmentation, the way we encode motion can resemble theirs with the key difference that they learn motion dynamics from real data and then draw samples from the learned data distribution; we, instead, learn a latent representation for motion trajectories by enforcing spatio-temporal consistency of generated content. Additionally, appearance and motion are also captured differently; for example,  
in \cite{pmlr-v80-jang18a}, they are included by using explicit values as conditions for the generation of future frames, namely, the first frame of the video and some motion attributes; we, instead, model motion and appearance simply as samples drawn from latent spaces and provided as inputs to the generator, and not as quantities estimated by the discriminator.}

As mentioned above, we adopt  GANs mainly for unsupervised learning of object motion. GANs have been already employed for unsupervised domain adaption~\cite{Bousmalis_2017_CVPR,Tzeng_2017_CVPR}, image-to-image translation~\cite{Zhu_2017_ICCV,Yi_2017_ICCV}, for semi-supervised semantic image segmentation~\cite{Souly_2017_ICCV}, \change{as well as for unsupervised feature learning in the image domain \cite{doersch2015unsupervised}}. In the video domain, GANs have been particularly useful for semi-supervised and unsupervised video action recognition~\cite{NIPS2016_6194,Saito_2017_ICCV,Tulyakov_2018_CVPR} or representation learning~\cite{Mahasseni_2017_CVPR}, given their innate ability to learn video dynamics while discriminating between real and fake videos. 
Unlike existing approaches, our video generation framework supports unsupervised pixel-wise dense prediction for video object segmentation, which is a more complex task that requires learning contextual relations between time-varying pixels.
To the best of our knowledge, this is the first attempt to perform  adversarial unsupervised video object segmentation, although some GAN-based approaches \cite{NIPS2017_6639} perform dense prediction for the optical flow estimation problem.
In particular, \cite{NIPS2017_6639} proposes a conditional GAN taking an image pair as input and predicting the optical flow. Flow-warped error images both on predicted flow and on ground-truth flow are then computed and a discriminator is trained to distinguish between the real and the fake ones. The network is trained both on labeled and unlabeled data and the adversarial GAN loss is extended with the supervised end-point-error loss, computed on the labeled data.
Differently from this work, our dense-prediction network uses only unlabeled data and extends the traditional adversarial loss by including the error made by the discriminator in estimating motion as well as in predicting segmentation maps.

\section{Adversarial Framework for Video Generation and Unsupervised Motion Learning}
Our adversarial framework for video generation and dense prediction --- \emph{VOS-GAN} --- is based on a GAN framework and consists of the following two modules:
\begin{itemize}
 \item a \emph{generator}, implemented as a hybrid deep CNN-RNN, that receives two inputs: 1) a noise vector from a latent space that models scene background; 2) a sequence of vectors that model foreground motion as a trajectory in another latent space. The output of the generator is a video with its corresponding foreground mask.
 \item a \emph{discriminator}, implemented as a deep CNN, that receives an input video and 1) predicts whether it is real or not; 2) performs pixel-wise dense prediction to generate an object segmentation map; 3) performs pixel-wise dense prediction to estimate the optical flow between video frames.
\end{itemize}
The traditional adversarial loss is extended by having the discriminator learn to compute motion-related dense predictions for the input video, thus forcing the generator to produce more realistic motion trajectories. Additionally, this formulation makes the discriminator suitable as a stand-alone model for object segmentation and optical flow estimation.

\label{sec:model}

\subsection{Generator Architecture}
\begin{figure*}[h]
\centering
\includegraphics[width=0.95\textwidth]{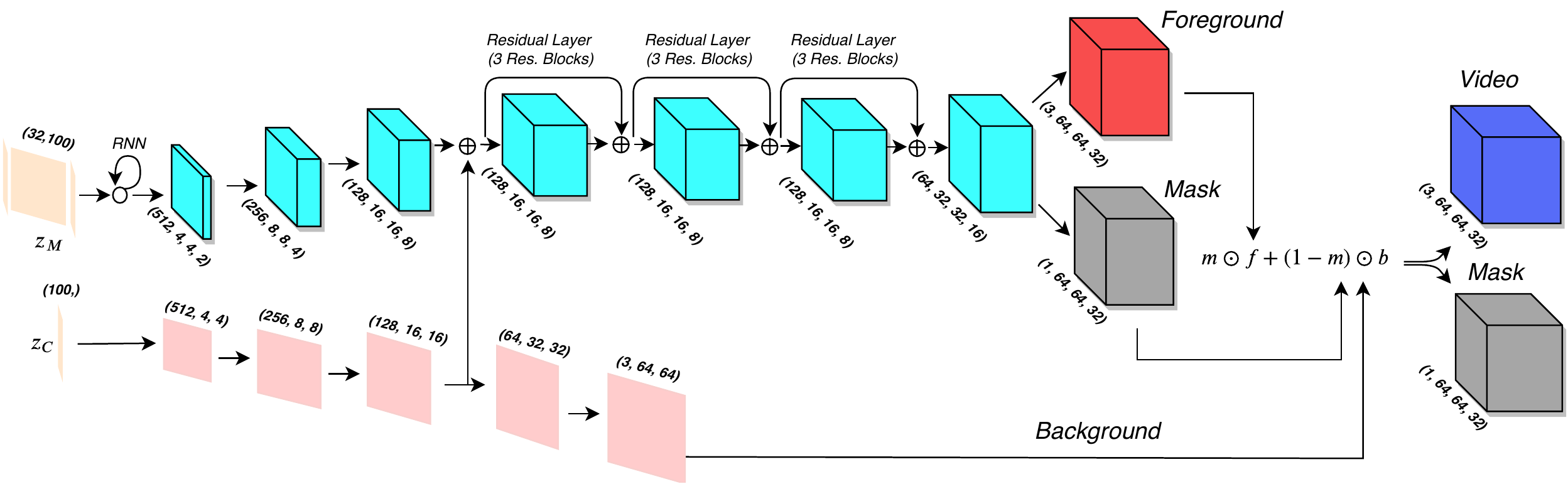}
\caption{\textbf{Generator architecture:} the \emph{background stream} (bottom) uses a latent vector defining the general scene of the video, and produces a background image; the \emph{foreground stream} (top) processes a sequence of latent vectors, obtained by spherically interpolating the start and end points, to generate frame-by-frame foreground appearance and motion masks. The foreground stream is conditioned on video content by concatenation (denoted with $\oplus$ in the figure) of intermediate features produced by the background stream. Information about dimensions of intermediate outputs is given in the figure in the format (\textit{channels, height, width, duration}) tuples.}
\label{fig:generator}
\end{figure*}

The architecture of the generator, inspired by the two-stream approach in~\cite{NIPS2016_6194}, is shown in Fig.~\ref{fig:generator}. Specifically, our generation approach factorizes the process into separate background and foreground generation, on the assumption that a scene is generally stationary and the presence of informative motion can be constrained only to a set of objects of interest in a semi-static environment. However, unlike \cite{NIPS2016_6194} and similar to \cite{Tulyakov_2018_CVPR}, we separate the latent spaces for scene and foreground generation, and explicitly represent the latter as a temporal quantity, thus enforcing a more natural correspondence between the latent input and the frame-by-frame motion output.

Hence, the generator receives two inputs: $z_C \in \mathcal{Z}_C = \mathbb{R}^d$ and $z_M = \{z_{M,i}\}_{i=1}^t$, with each $z_{M,i} \in \mathcal{Z}_M = \mathbb{R}^d$. A point $z_C$ in the latent space $\mathcal{Z}_C$ encodes the general scene to be applied to the output video, and is mainly responsible for driving the \emph{background stream} of the model. This stream consists of a cascade of transposed convolutions, which gradually increase the spatial dimension of the input in order to obtain a full-scale background image $b(z_C)$, that is used for all frames in the generated video.

The set of $z_{M,i}$ points from the latent space, $\mathcal{Z}_M$, defines the objects motion to be applied in the video. The latent sequence is obtained by sampling the initial and final points and performing a spherical linear interpolation (SLERP~\cite{Shoemake:1985}) to compute all intermediate vectors, such that the length of the sequence is equal to the length (in frames) of the generated video. Using an interpolation rather than sampling multiple random points enforces temporal coherency in the foreground latent space.
The list of latent points is then encoded through a recurrent neural network (LSTM) in order to provide a single vector (i.e., the LSTM's final state) summarizing a representation of the whole motion. %The objective of using a recurrent layer to process initially the latent trajectory is to make the whole process independent from the interpolation method to represent latent trajectories.
After a cascade of spatio-temporal convolutions (i.e., with 3D kernels that also span the time dimension), motion features are conditioned on content by concatenation of intermediate features from the background that are replicated along the time dimension. Then, these activations are processed by a $1\times1\times1$ convolution and \change{three 3D residual layers \cite{he2016deep}, with three residual blocks in each layer}. After another convolutional layer, this \textit{foreground stream} outputs a set of frames $f(z_C, z_M)$ with foreground content and masks defining motion pixel location $m(z_C, z_M)$.

The two streams are finally combined as
\begin{equation}
\begin{split}
G(z_C, z_M) = m(z_C, z_M) \odot f(z_C, z_M) \\ + (1 - m(z_C, z_M)) \odot b(z_C)
\end{split}
\end{equation}

Foreground generation can be directly controlled acting on $z_M$. Indeed, varying $z_M$ for a fixed value of $z_C$ results in videos with the same background and different foreground appearance and motion.

\subsection{Discriminator Architecture}
The primary goal of the discriminator network is to distinguish between generated and real videos, in order to push the generator towards more realistic outputs. At the same time, we train the discriminator to perform dense pixel-wise predictions of foreground masks and optical flow. These two additional outputs have a twofold objective: 1) they force the discriminator to learn motion-related features, rather than (for example) learn to identify the visual features of objects that are more likely to be part of the foreground (e.g., people, animals, vehicles); 2) they enable the discriminator to perform additional tasks from unlabeled data.

\begin{figure*}[h]
\centering
\includegraphics[width=0.95\textwidth]{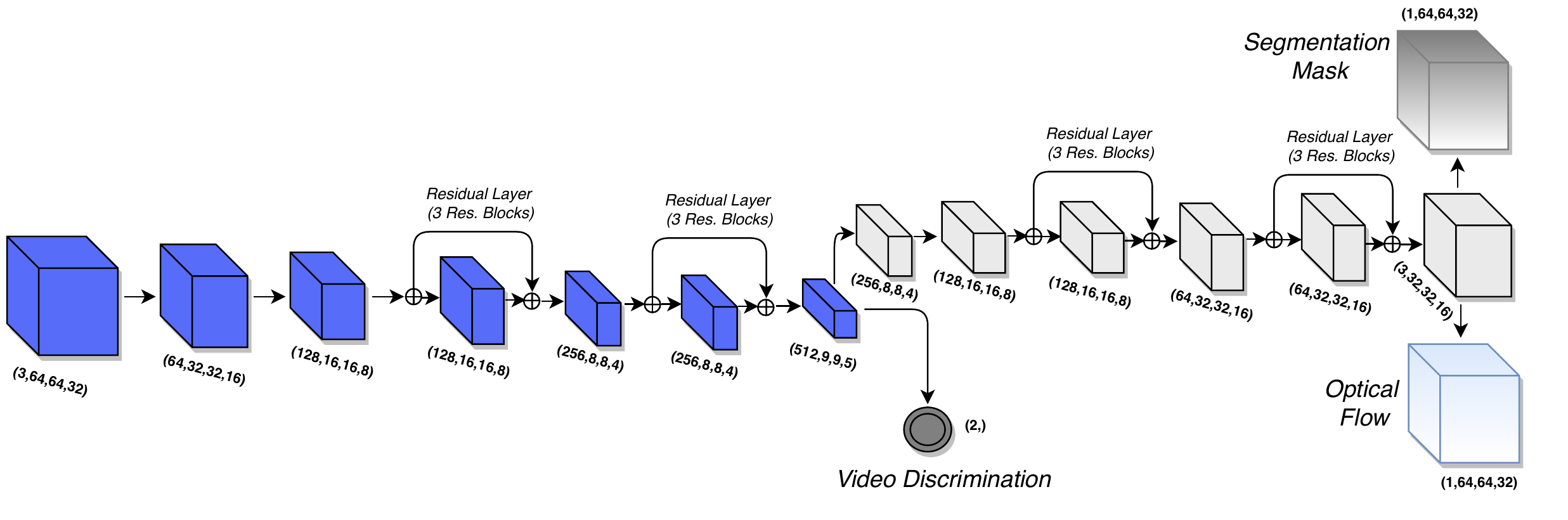}
\caption{\textbf{Discriminator architecture:} the \emph{motion stream} (top) predicts foreground map and optical flow of the input video; the \emph{discrimination stream} (bottom) outputs a single value, used for adversarial training, predicting whether the input video is real or fake.}
\label{fig:placeholder_discr}
\end{figure*}

\change{Fig.~\ref{fig:placeholder_discr} shows the architecture of the discriminator. The input to the model is a video clip (either real or produced by the generator), that goes first through a series of convolutional and residual layers, encoding the video dynamics into a more compact representation, which in turn is provided as input to two separate streams: 1) a \emph{discrimination stream}, which applies 3D convolutions to the intermediate representation and then makes a prediction on whether the input video is real or fake; 2) a \emph{motion stream}, feeding the intermediate representation to a cascade of 3D transposed convolutional and residual layers, which fork at the final layer and return the frame-by-frame foreground segmentation maps (each as a 2D binary map) and optical flow estimations (each as a two-channel 2D map) for the input video.}

The discrimination path of the model (i.e., the initial shared convolutional layers and the discrimination stream) follows a standard architecture for video discrimination \cite{NIPS2016_6194}, while the motion path, based on transposed convolutions, decodes the video representation in order to provide the two types of dense predictions \cite{long2015fully}. Formally, we define the outputs returned by the discriminator as $D_{\text{adv}}(x)$, $D_{\text{seg}}(x)$ and $D_{\text{opt}}(x)$, which are, respectively, the probability that the provided input is real, the foreground segmentation maps and the optical flow estimated for the video; input $x$ may be either a real video or the output of the generator.

\subsection{Learning Procedure}
We jointly train the generator and the discriminator in a GAN framework, with the former trying to maximize the probability that the discriminator predicts fake outputs as real, and the latter trying to minimize the same probability. Additionally, when training the discriminator, we also include loss terms related to the accuracy of the estimated foreground segmentation and optical flow.

The main problem in computing these additional losses is that, while optical flow ground truth for fake videos can be easily obtained by assuming the output of a state-of-the-art algorithm to be sufficiently accurate, there are no video segmentation approaches that provide the needed accuracy. To solve this problem, we propose what is --- to the best of our knowledge --- the first approach for video object segmentation trained in an unsupervised manner: since the architecture of the generator internally computes the foreground masks of the generated videos, we use those to supervise the prediction of the masks computed by the discriminator. Of course, this kind of self-referential ground-truth will not be very meaningful at first, but will become more and more akin to real ground-truth from real videos as the generator improves.

The discriminator loss is then defined as follows (for the sake of compactness, we will define $z = (z_C, z_M)$):
%\begin{equation}
%\begin{split}
%\mathcal{L}_D = -\mathbb{E}_{x \sim p_\text{real}}\left[ \log D_\text{adv}\left(x\right) \right]
%- \mathbb{E}_{x \sim p_z}\left[ \log \left( 1 - D_\text{adv}\left( G\left(z\right) \right) \right) \right] \\
%+ \mathbb{E}_{z \sim p_z}\left[ \frac{1}{n^2} \sum_{i,j} \text{NLL}\left(D_\text{seg}\left(G\left(z\right)\right)_{(i,j)}, m(z)_{(i,j)}\right) \right] \\
%+ \mathbb{E}_{z \sim p_z}\left[ \frac{1}{n^2} \sum_{i,j} \left(D_\text{opt}\left(G\left(z\right)\right)_{(i,j)} - \text{OF}\left(G\left(z\right)\right)_{(i,j)}\right)^2 \right]\\
%+ \mathbb{E}_{x \sim p_\text{real}}\left[ \frac{1}{n^2} \sum_{i,j} \left(D_\text{opt}(x_{(i,j)}) - \text{OF}\left(x\right)_{(i,j)}\right)^2 \right] 
%\end{split}
%\end{equation}

\begin{equation}\label{eq:loss_D}
\begin{split}
\mathcal{L}_D =& -\mathbb{E}_{x \sim p_\text{real}}\left[ \log D_\text{adv}\left(x\right) \right] \\ &
- \mathbb{E}_{z \sim p_z}\left[ \log \left( 1 - D_\text{adv}\left( G\left(z\right) \right) \right) \right] +\\
& + \mathbb{E}_{z \sim p_z}\left[  \text{NLL}_\text{2D}\left(D_\text{seg}\left(G\left(z\right)\right), m(z)\right) \right] +\\
& + \alpha \left(\mathbb{E}_{z \sim p_z}\left[\left\|D_\text{opt}\left(G\left(z\right)\right)-\text{OF}\left(G\left(z\right)\right)\right\|^2 \right] \right) +\\
& + \alpha \left(\mathbb{E}_{x \sim p_\text{real}}\left[ \left\|D_\text{opt}\left(x\right) - \text{OF}\left(x\right)\right\| ^2 \right] \right)
\end{split}
\end{equation}

In the equation above, the first two lines encode the adversarial loss, which pushes the discriminator to return high likelihood scores for real videos and low ones for the generated videos. The third line encodes the loss on foreground segmentation, and it computes the average pixel-wise negative log-likelihood of the predicted segmentation map, using the generator's foreground mask $m(z)$ as source for correct labels (in our notation, $\text{NLL}_\text{2D}(\hat{y}, y)$ is the negative log-likelihood of predicted label map $\hat{y}$ given correct label map $y$). The last two lines encode the loss on optical flow estimation, as the squared $L_2$ norm between the predicted optical flow and the one calculated on the input video using the $\text{OF}(\cdot)$ function, implemented as per \cite{Farneback:2003}. It should be noted that the non-adversarial term related to object segmentation is only computed on the generated videos (hence in a fully-unsupervised manner) for which foreground masks are provided by the generator in Fig. \ref{fig:generator}, since segmentation ground-truth may not be available for real videos. Optical flow is, instead, computed on both generated and real videos, and it serves to provide supervision to the discriminator (especially to the motion stream shown in Fig. \ref{fig:placeholder_discr}) in learning motion cues from real videos.
\change{We additionally introduce an $\alpha$ term to control the influence of optical flow and segmentation estimations over the traditional discriminator loss. $\alpha$ is initially set to 1 and then increased by a step $s_{\alpha}$ at each epoch since a specific epoch $E_{\alpha}$. The role of $\alpha$ is specifically to stabilize the GAN training procedure by first letting it learn the general appearance of the scene and, as training goes on, focusing more on learning motion cues.}

The generator loss is, more traditionally, defined as:
\begin{equation}
 \mathcal{L}_G = -\mathbb{E}_{z \sim p_z}\left[ \log D_\text{adv}\left(G\left( z \right) \right) \right]
\end{equation}
In this case, the generator tries to push the discriminator to increase the likelihood of its output being real.

During training, we follow the common approach for GAN training, by sampling real videos (from an existing dataset) and generated videos (from the generator) and alternately optimizing the discriminator and the generator.

\section{Performance Analysis}
Our system is specifically designed for supporting unsupervised dense and global prediction in videos, but it requires first to train the GAN-based generation model. For this reason, we initially report  the video generation performance followed by the performance obtained for video object  segmentation (pixel-wise dense prediction) and video action recognition (global prediction).
\subsection{Datasets and Metrics} 
\subsubsection{Video Generation}

For ease of comparison with publicly-available implementations of state-of-the-art methods, we train our video generation model on two different datasets: ``Golf course'' videos (over 600,000 video clips, each 32 frames long) from the dataset proposed in \cite{NIPS2016_6194} and also employed in \cite{Saito_2017_ICCV}, and the Weizmann Action database \cite{gorelick2007} (93 videos of people performing 9 actions), employed in \cite{Tulyakov_2018_CVPR}. To have data in the same format, we pre-process the Weizmann Action dataset by splitting the original videos into 32-frame partially-overlapping (i.e., shifted by 5 frames) sequences, resulting in 683 such video clips.

For testing video generation capabilities, we compute qualitative and quantitative results. For the former, we carry out a user study aiming at assessing how the generated videos are perceived by humans in comparison to other GAN-based generative methods, and measure the preference score in percentage.
To assess video generation performance quantitatively, we evaluate the appearance and motion of the generated videos, using the following  metrics:
\begin{itemize}
	\item {\bf Foreground Content Distance (FCD)}. This score assesses foreground appearance consistency by computing the average $L_2$ distance between visual features of foreground objects in two consecutive frames. Feature vectors are extracted from a fully-connected layer of a pre-trained Inception network \cite{43022}, whose input is the bounding box containing the foreground region, defined as the discriminator's segmentation output.
	%\item {\bf Motion coherency (MC)}. While the previous score describes the quality of the generated visual appearance of moving objects, we propose a metric to evaluate how realistic the generated motion is, by computing the KL-divergence between magnitude/orientation histograms of optical flows of real and generated videos.
	\item \change{\textbf{Fr\'echet Inception Distance (FID)} \cite{NIPS2017_7240} suitably adapted to videos as in \cite{wang2018vid2vid}. FID is a widely adopted metric for implicit generative models, as it has been demonstrated to correlate well with visual quality of generated content. We employ the variant for videos proposed in \cite{wang2018vid2vid}, that projects generated videos into a feature space corresponding to a specific layer of a pre-trained video recognition CNN. The embedding layer is then considered as a continuous multivariate Gaussian, and consequently mean and covariance are computed for both generated data and real data. The Fr\'echet distance between these two Gaussians (i.e., Wasserstein-2 distance) is then used to quantify the quality of generated samples. As pre-trained CNN model for feature extraction we use 
	%I3D \cite{8099985}.} 
	ResNeXt~\cite{xie2017aggregated,hara2018can}.}
	\item {\bf Inception score (IS)} \cite{NIPS2016_6125} is the most adopted quality score in GAN literature. In our case, we compute the Inception score by sampling a random frame from each video in a pool of generated ones.
\end{itemize}

\subsubsection{Video Object Segmentation}
The capabilities of the motion stream of our discriminator network (see Fig. \ref{fig:placeholder_discr}) for pixel-wise dense prediction are tested on \change{several benchmarks for video object segmentation: DAVIS 2016 and DAVIS 2017 \cite{7780454} and SegTrack-v2 \cite{Tsai2012}. Each dataset provides accurate pixel-level annotations for all video frames. The employed datasets show diverse features, useful to test the performance of video object segmentation methods in different scenarios. In particular, DAVIS 2017 contains 150 video sequences (50 of which constitute the DAVIS 2016 dataset), and includes challenging examples with occlusion, motion blur and appearance changes.
SegTrack-v2 is a dataset containing 14 videos showing camera motion, slow object motion, object/background similarity, non-rigid deformations and articulated objects.} %F4K-Fish contains 17 videos taken in an unconstrained underwater scenario and is characterized by small objects, highly-cluttered scenes with occlusions and illumination changes.}

We employ standard metrics \cite{7780454} for measuring video object segmentation performance to ease comparison to state-of-the-art methods, namely: a) \textit{region similarity} $\mathcal{J}$, computed as pixel-wise intersection over union of the estimated segmentation and the ground-truth mask; b) \textit{contour accuracy $\mathcal{F}$}, defined as the F$_1$-measure between the contour points of the estimated segmentation mask and the ground-truth mask. 
For each of the two metrics, we compute the mean value as:
\begin{equation}
\mathcal{M}_\mathcal{C}(R)= \frac{1}{|R|} \sum_{\mathcal{S}_i \in \mathcal{R}} \overline{\mathcal{C}}(S_i),
\end{equation}
where $\overline{\mathcal{C}}(S_i)$ is the average of measure $\mathcal{C}$ (either $\mathcal{J}$ or  $\mathcal{F}$) on $S_i$ and $R$ is the set of video sequences $S$. Also, for comparison with state of the art methods we compute recall $\mathcal{O}$ and decay $\mathcal{D}$ of the above metrics. The former quantifies the portion of sequences on which a segmentation method scores higher than a threshold, and is defined as:
\begin{equation}
\mathcal{O}_\mathcal{C} (R)= \frac{1}{|R|} \sum_{\mathcal{S}_i \in \mathcal{R}} \mathbb{I}\left[
\overline{\mathcal{C}}(S_i)> \tau\right],
\end{equation}
where $\mathbb{I}(p)$ is 1 if $p$ is true and 0 otherwise, and $\tau$ = 0.5 in our experiments. Decay measures the performance loss (or gain) over time:
\begin{equation}
\mathcal{D}_\mathcal{C} (R)= \frac{1}{|R|} \sum_{\mathcal{Q}_i \in \mathcal{R}} \overline{\mathcal{C}}(Q_i^1) - \overline{\mathcal{C}}(Q_i^4), 
\end{equation}
with $Q_i = \{ Q_i^1, Q_i^2, Q_i^3, Q_i^4 \}$ being a partition of $S_i$ in quartiles.

\change{The results are computed on test or validation sets where available; otherwise, we split the videos into training and test sets with proportions of 70\% and 30\%. 
In particular, ablation studies and analysis of the different architectural settings are done on DAVIS 2017 (because of its larger size). We use DAVIS 2016 and SegTrack-v2 for showing generalization capabilities of our approach and for comparison to state-of-the-art methods.}
All available videos are divided into 32-frame shots and downsampled to fit the input size allowed by the video segmentation network (i.e., the discriminator architecture in Fig. \ref{fig:placeholder_discr}), i.e., 64$\times$64, while output segmentation maps are rescaled (through bi-linear interpolation) to ground-truth size for metrics computation.

Accurate evaluation of optical flow estimation is not performed, since our model is designed primarily for performing prediction by self-supervision (i.e., adversarial generation of foreground masks), while optical flow, provided by a state-of-the-art method, is used only to guide the discriminator towards learning motion features from real videos.
\change{It should be noted that we did not use any deep learning--based optical flow, e.g., \cite{IMSKDB17}, but, instead, employed the traditional approach in \cite{Farneback:2003} as it it exploits physical properties of object motion in a purely unsupervised way. This avoids to include any form of ``human-supervision'' in the segmentation pipeline, making the proposed approach fully unsupervised}.

\subsubsection{Action Recognition}
The capabilities of our model (namely, the discriminator stream in Fig. \ref{fig:placeholder_discr}) in learning global motion cues are tested in a video action recognition scenario. In particular, we evaluate its accuracy in classifying actions in videos on on two benchmarks, UCF101 \cite{ucf101} (13,320 videos from 101 action categories) and the Weizmann Action database \cite{gorelick2007}. We use average classification accuracy (the ratio of correctly-classified samples over the total number of test samples) as metric. In the evaluation on Weizmann Action dataset, due to the small number of video sequences, we perform 10-fold cross validation to average accuracy scores.
Performance on video action recognition serves also to provide an additional metrics for quantitative evaluation for video generation.

%Finally, it should be noted that we do not aim at performing an evaluation of the accuracy of our discriminator as an optical flow estimator. We only employ optical flow as a means to guide the discriminator towards learning meaningful motion features; however, this part of our evaluation focuses on video object segmentation, both unsupervised and supervised (via fine-tuning), as a practical use case for applying adversarial models.% Measuring the accuracy of the estimated optical flow would not be unsupervised, as the model is shown ``correct'' (though algorithm-based, but we consider them as good as ground-truth for our purposes) optical flow maps during training. 
\begin{table*}[h!]
\centering
\begin{tabular}{lcccccc}
\toprule
\textbf{Bkg stream} & Kernel & Stride & Padding & Activation & BatchNorm & Output shape \\
\midrule[0.001pt] 
$z_C$ & $-$ & $-$ & $-$ & $-$ & - & 100x1x1 \\
ConvTran2D & 4x4 & 1x1 & $-$ & LReLU($\alpha=0.2$) & Yes & 512x4x4 \\
ConvTran2D & 4x4 & 2x2 & 1x1 & LReLU($\alpha=0.2$) & Yes & 256x8x8 \\
ConvTran2D & 4x4 & 2x2 & 1x1 & LReLU($\alpha=0.2$) & Yes & 128x16x16 \\
ConvTran2D & 4x4 & 2x2 & 1x1 & LReLU($\alpha=0.2$) & Yes & 64x32x32 \\
ConvTran2D & 4x4 & 2x2 & 1x1 & Tanh & No & 3x64x64 \\
\bottomrule
\end{tabular}

\begin{tabular}{lcccccc} \toprule
\textbf{Motion features} & Kernel Size & Stride & Padding & Activation & BatchNorm & Output shape \\
\midrule[0.001pt] 
$\text{RNN}(z_M)$ & $-$ & $-$ & $-$ & $-$ & No & 100x1x1 \\
ConvTran3D & 2x4x4 & 1x1x1 & $-$ & LReLU($\alpha=0.2$) & Yes & 512x4x4x2 \\
ConvTran3D & 3x3x3 & 3x3x3 & 1x2x2 & LReLU($\alpha=0.2$) & Yes & 256x8x8x4 \\
ConvTran3D & 4x4x4 & 2x2x2 & 1x1x1 & LReLU($\alpha=0.2$) & Yes & 128x16x16x8 \\
\end{tabular}

\begin{tabular}{lcccccc} \toprule
\textbf{Fg features} & Kernel Size & Stride & Padding & Activation & BatchNorm & Output shape \\
\midrule[0.001pt]
Conv3D & 1x1x1 & 1x1x1 & $-$ & $-$ & $-$ & 128x16x16x8 \\
\midrule[0.001pt]
\multicolumn{7}{l}{Residual Layer}\\
\midrule[0.001pt]
3D Residual Block & 3x3x3 & 1x1x1 & 1x1x1 & ReLU & Yes & 128x16x16x8 \\
3D Residual Block & 3x3x3 & 1x1x1 & 1x1x1 & ReLU & Yes & 128x16x16x8 \\
3D Residual Block & 3x3x3 & 1x1x1 & 1x1x1 & ReLU & Yes & 128x16x16x8 \\
\midrule[0.001pt]
\multicolumn{7}{l}{Residual Layer}\\
\midrule[0.001pt]
3D Residual Block & 3x3x3 & 1x1x1 & 1x1x1 & ReLU & Yes & 128x16x16x8 \\
3D Residual Block & 3x3x3 & 1x1x1 & 1x1x1 & ReLU & Yes & 128x16x16x8 \\
3D Residual Block & 3x3x3 & 1x1x1 & 1x1x1 & ReLU & Yes & 128x16x16x8 \\
\midrule[0.001pt]
\multicolumn{7}{l}{Residual Layer}\\
\midrule[0.001pt]
3D Residual Block & 3x3x3 & 1x1x1 & 1x1x1 & ReLU & No & 128x16x16x8 \\
3D Residual Block & 3x3x3 & 1x1x1 & 1x1x1 & ReLU & No & 128x16x16x8 \\
3D Residual Block & 3x3x3 & 1x1x1 & 1x1x1 & ReLU & No & 128x16x16x8 \\
\midrule[0.001pt]
ConvTran3D & 4x4x4 & 2x2x2 & 1x1x1 & LReLU($\alpha=0.2$) & Yes & 64x32x32x16 \\
\end{tabular}

\begin{tabular}{lcccccc} \toprule
\textbf{Fg raw} & Kernel Size & Stride & Padding & Activation & BatchNorm & Output shape \\
\midrule[0.001pt]
ConvTran3D & 4x4x4 & 2x2x2 & 1x1x1 & Tanh & No & 3x64x64x32 \\ 
\end{tabular}
%\midrule
\begin{tabular}{lcccccc} \toprule
\textbf{Fg mask} & Kernel Size & Stride & Padding & Activation & BatchNorm & Output shape \\
\midrule[0.001pt] 
ConvTran3D & 4x4x4 & 2x2x2 & 1x1x1 & Sigmoid & No & 1x64x64x32 \\ 
\bottomrule
\end{tabular}
\quad \\
	\caption{Architecture of the generator. \textit{Bkg stream} contains the layers included in the background stream of the model, that returns $b(z_C)$ (see Eq. 1 in the paper). \textit{Motion features} lists the layers in the initial shared part of the foreground stream. \textit{Fg features} considers the layers used to process concatenation over the channel dimension of the output of \textit{motion features} with the output from the third layer of the \textit{background stream}. By "3D Residual Block" we simply denote standard shape-preserving residual blocks using 3D convolution. The output of \textit{Fg raw} is $f(z_C, z_M)$ and the output of {Fg mask} is $m(z_C, z_M)$. \textit{LReLU} stands for Leaky ReLU, while layers marked with \textit{ConvTran} execute transposed convolution over their input.} \label{tab:generator}
\end{table*}

\begin{table*}[h!]
\centering
\begin{tabular}{ccccccc} \toprule
\textbf{Shared} & Kernel Size & Stride & Padding & Activation & BatchNorm & Out shape \\
\midrule[0.001pt] 
Input & $-$ & $-$ & $-$ & $-$ & & 3x64x64x32\\
Conv3D & 4x4x4 & 2x2x2 & 1x1x1 & LReLU($\alpha = 0.2$) & No & 64x32x32x16\\
Conv3D & 4x4x4 & 2x2x2 & 1x1x1 & LReLU($\alpha = 0.2$) & Yes & 128x16x16x8\\
\midrule[0.001pt] 
\multicolumn{7}{l}{Residual Layer}\\
\midrule[0.001pt] 
3D Residual Block & 3x3x3 & 1x1x1 & 1x1x1 & ReLU & Yes & 128x16x16x8 \\
3D Residual Block & 3x3x3 & 1x1x1 & 1x1x1 & ReLU & Yes& 128x16x16x8 \\
3D Residual Block & 3x3x3 & 1x1x1 & 1x1x1 & ReLU & Yes& 128x16x16x8 \\
\midrule[0.001pt] 
Conv3D & 4x4x4 & 2x2x2 & 1x1x1 & LReLU($\alpha = 0.2$) & Yes & 256x8x8x4\\
\midrule[0.001pt] 
\multicolumn{7}{l}{Residual Layer}\\
\midrule[0.001pt] 
3D Residual Block & 3x3x3 & 1x1x1 & 1x1x1 & ReLU & Yes & 256x8x8x4 \\
3D Residual Block & 3x3x3 & 1x1x1 & 1x1x1 & ReLU & Yes & 256x8x8x4 \\
3D Residual Block & 3x3x3 & 1x1x1 & 1x1x1 & ReLU & Yes&  256x8x8x4 \\
\midrule[0.001pt] 
Conv3D & 2x2x2 & 1x1x1 & 1x1x1 & LReLU($\alpha = 0.2$) & Yes & 512x9x9x5\\
\bottomrule
\end{tabular}

\begin{tabular}{ccccccc} \toprule
\textbf{Motion} & Kernel Size & Stride & Padding & Activation & BatchNorm & Out shape \\
\midrule[0.001pt] 
ConvTran3D & 2x2x2 & 1x1x1 & 1x1x1 & ReLU & Yes & 256x8x8x4\\
ConvTran3D & 4x4x4 & 2x2x2 & 1x1x1 & ReLU & Yes & 128x16x16x8\\
\midrule[0.001pt] 
\multicolumn{7}{l}{Residual Layer}\\
\midrule[0.001pt] 
3D Residual Block & 3x3x3 & 1x1x1 & 1x1x1 & ReLU & Yes & 128x16x16x8 \\
3D Residual Block & 3x3x3 & 1x1x1 & 1x1x1 & ReLU & Yes& 128x16x16x8 \\
3D Residual Block & 3x3x3 & 1x1x1 & 1x1x1 & ReLU & Yes& 128x16x16x8 \\
\midrule[0.001pt] 
ConvTran3D & 4x4x4 & 2x2x2 & 1x1x1 & ReLU & Yes & 64x32x32x16\\
\midrule[0.001pt] 
\multicolumn{7}{l}{Residual Layer}\\
\midrule[0.001pt] 
3D Residual Block & 3x3x3 & 1x1x1 & 1x1x1 & ReLU & Yes & 64x32x32x16 \\
3D Residual Block & 3x3x3 & 1x1x1 & 1x1x1 & ReLU & Yes& 64x32x32x16 \\
3D Residual Block & 3x3x3 & 1x1x1 & 1x1x1 & ReLU & Yes & 64x32x32x16 \\
\midrule[0.001pt] 
ConvTran3D & 4x4x4 & 2x2x2 & 1x1x1 & Sigmoid & No & 3x64x64x32\\
\bottomrule
\end{tabular}

\begin{tabular}{ccccccc} \toprule
\textbf{Discr.} & Kernel Size & Stride & Padding & Activation & BatchNorm & Out shape \\
\midrule[0.001pt] 
Conv3D & 4x4x4 & 2x2x2 & 1x1x1 & LReLU($\alpha=0.2$) & Yes & 1024x4x4x2 \\
Conv3D & 4x4x4 & 4x4x2 & $-$ & Softmax & No & 2x1x1x1 \\
\bottomrule
\end{tabular}
\caption{Discriminator architecture. Note that the structure of transposed convolutions used in \textit{Motion Stream} is symmetrical to the one of convolutions in the \textit{Shared} part. \textit{LReLU} stands for Leaky ReLU, while the layers marked with \textit{ConvTran} are transposed convolutions. The output of the \textit{motion stream} has 3 channels: 1 for segmentation and 2 for optical flow.}\label{tab:discriminator}.
\end{table*}

\subsection{Training Settings}
The architectures of the generator and discriminator networks in terms of kernel sizes, padding, stride, activation functions and use of batch normalization \cite{ioffe2015batch} are given, respectively, in Tables \ref{tab:generator} and \ref{tab:discriminator}.

In the video generation and segmentation experiments, we performed gradient-descent using ADAM, with an initial learning rate of 0.0002, $\beta_1 = 0.5$, $\beta_2 = 0.999$ and batch size of 16. For video generator training we used $\alpha=1$, $s_{\alpha}=0.2$ and $E_{\alpha}=2$ (values set empirically). 
We trained the video generation models for 25 epochs and the video segmentation ones for 200 epochs. For video action recognition, we used SGD with momentum and dampening of 0.9, weight decay of 0.001 and learning rate of 0.1, reduced by a factor of 10 when no improvement in validation accuracy occurred for 10 epochs. Batch size was 128 and the number of epochs was 130. %Our code, as well as dataset splits and pre-trained models, will be publicly released.

\subsection{Video Generation}
\subsubsection{Qualitative evaluation}
To evaluate qualitatively our video generation approach, we used Amazon Mechanical Turk (MTurk) in order to measure how generated videos are perceived by humans. Our generated videos are compared to those synthesized by VGAN~\cite{NIPS2016_6194}, TGAN~\cite{Saito_2017_ICCV}, MoCoGAN~\cite{Tulyakov_2018_CVPR}.
%Each of the compared models was first trained on the ``golf course'' videos, then two additional variants were obtained by fine-tuning the trained GAN on videos from either UCF101 or DAVIS. On MTurk, each job was created by randomly choosing two of the models under comparison and generating, for each, three 16-video batches from the three trained instances of that model; workers were then asked to choose which of the two sets of videos looked more realistic.
%TGAN and VGAN are trained (using original code) on the ``golf course'' videos, while MoCoGAN is trained on the Weizmann Action database. 
%We compare these models to variants of our VOS-GAN trained on each of the two datasets, with the resulting models indicated in the following as \emph{VOS-GAN\textsubscript{G}} (trained on ``golf-course'') and \emph{VOS-GAN\textsubscript{W}} (trained on Weizmann Action dataset). 

\begin{figure*}[h]
	\centering
	%\textbf{Golf Course}\\ \vspace{0.1cm}
	\includegraphics[width=0.06\textwidth]{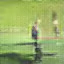}
	\includegraphics[width=0.06\textwidth]{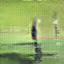}
	\includegraphics[width=0.06\textwidth]{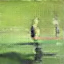}
	\includegraphics[width=0.06\textwidth]{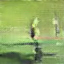}
	\includegraphics[width=0.06\textwidth]{frames/vgan/video1/32} \hspace{0.03cm}
	\includegraphics[width=0.06\textwidth]{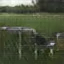}
	\includegraphics[width=0.06\textwidth]{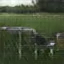}
	\includegraphics[width=0.06\textwidth]{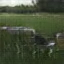}
	\includegraphics[width=0.06\textwidth]{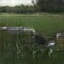}
	\includegraphics[width=0.06\textwidth]{frames/vgan/video2/32} \hspace{0.03cm}
	\includegraphics[width=0.06\textwidth]{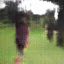}
	\includegraphics[width=0.06\textwidth]{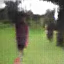}
	\includegraphics[width=0.06\textwidth]{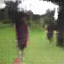}
	\includegraphics[width=0.06\textwidth]{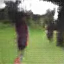}
	\includegraphics[width=0.06\textwidth]{frames/vgan/video3/32} \\ \vspace{0.4cm}
	\includegraphics[width=0.06\textwidth]{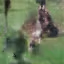}
	\includegraphics[width=0.06\textwidth]{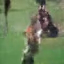}
	\includegraphics[width=0.06\textwidth]{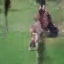}	
	\includegraphics[width=0.06\textwidth]{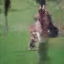}
	\includegraphics[width=0.06\textwidth]{frames/tgan/video1/32} \hspace{0.03cm}
	\includegraphics[width=0.06\textwidth]{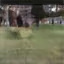}
	\includegraphics[width=0.06\textwidth]{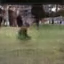}
	\includegraphics[width=0.06\textwidth]{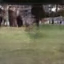}
	\includegraphics[width=0.06\textwidth]{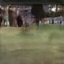}
	\includegraphics[width=0.06\textwidth]{frames/tgan/video2/32} \hspace{0.03cm}
	\includegraphics[width=0.06\textwidth]{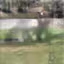}
	\includegraphics[width=0.06\textwidth]{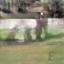}
	\includegraphics[width=0.06\textwidth]{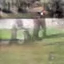}
	\includegraphics[width=0.06\textwidth]{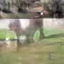}
	\includegraphics[width=0.06\textwidth]{frames/tgan/video3/32} \\ \vspace{0.4cm}
	\includegraphics[width=0.06\textwidth]{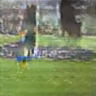}
	\includegraphics[width=0.06\textwidth]{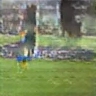}
	\includegraphics[width=0.06\textwidth]{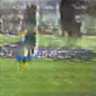}
	\includegraphics[width=0.06\textwidth]{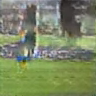}
	\includegraphics[width=0.06\textwidth]{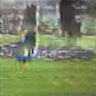} \hspace{0.03cm}
	\includegraphics[width=0.06\textwidth]{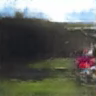}
	\includegraphics[width=0.06\textwidth]{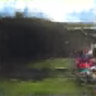}
	\includegraphics[width=0.06\textwidth]{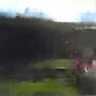}
	\includegraphics[width=0.06\textwidth]{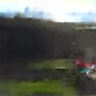}
	\includegraphics[width=0.06\textwidth]{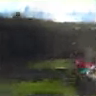} \hspace{0.03cm}
	\includegraphics[width=0.06\textwidth]{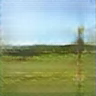}
	\includegraphics[width=0.06\textwidth]{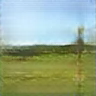}
	\includegraphics[width=0.06\textwidth]{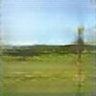}
	\includegraphics[width=0.06\textwidth]{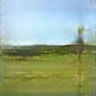}
	\includegraphics[width=0.06\textwidth]{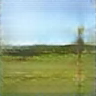}\\ \vspace{0.4cm}
	\includegraphics[width=0.06\textwidth]{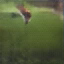}
	\includegraphics[width=0.06\textwidth]{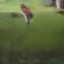}
	\includegraphics[width=0.06\textwidth]{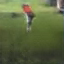}
	\includegraphics[width=0.06\textwidth]{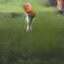}
	\includegraphics[width=0.06\textwidth]{frames/vos/1/32} \hspace{0.03cm}
	\includegraphics[width=0.06\textwidth]{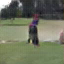}
	\includegraphics[width=0.06\textwidth]{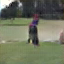}
	\includegraphics[width=0.06\textwidth]{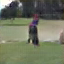}
	\includegraphics[width=0.06\textwidth]{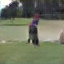}
	\includegraphics[width=0.06\textwidth]{frames/vos/6/32} \hspace{0.03cm}
	\includegraphics[width=0.06\textwidth]{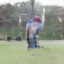}
	\includegraphics[width=0.06\textwidth]{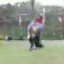}
	\includegraphics[width=0.06\textwidth]{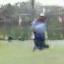}
	\includegraphics[width=0.06\textwidth]{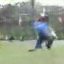}
	\includegraphics[width=0.06\textwidth]{frames/vos/16/32}
	\caption{\textbf{Frame samples on ``Golf course''}. (First row) VGAN-generated video frames show very little object motion, while (second row) TGAN-generated video frames show motion, but the quality of foreground appearance is low. (Third row)  MoCoGAN-generated videos: background quality is high, but there is little object motion.
	VOS-GAN (fourth row) generates video frames with a good compromise between object motion and appearance.}
	\label{fig:samples_golf}
\end{figure*}

\begin{figure*}[h]
	\centering
	%\textbf{Golf Course}\\ \vspace{0.1cm}
	\includegraphics[width=0.06\textwidth]{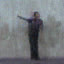}
	\includegraphics[width=0.06\textwidth]{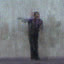}
	\includegraphics[width=0.06\textwidth]{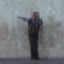}
	\includegraphics[width=0.06\textwidth]{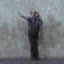}
	\includegraphics[width=0.06\textwidth]{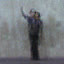} \hspace{0.03cm}
	\includegraphics[width=0.06\textwidth]{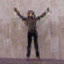}
	\includegraphics[width=0.06\textwidth]{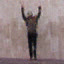}
	\includegraphics[width=0.06\textwidth]{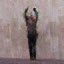}
	\includegraphics[width=0.06\textwidth]{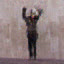}
	\includegraphics[width=0.06\textwidth]{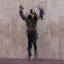} \hspace{0.03cm}
	\includegraphics[width=0.06\textwidth]{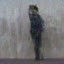}
	\includegraphics[width=0.06\textwidth]{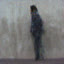}
	\includegraphics[width=0.06\textwidth]{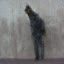}
	\includegraphics[width=0.06\textwidth]{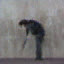}
	\includegraphics[width=0.06\textwidth]{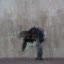} \\ \vspace{0.4cm}
	\includegraphics[width=0.06\textwidth]{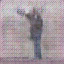}
	\includegraphics[width=0.06\textwidth]{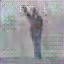}
	\includegraphics[width=0.06\textwidth]{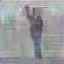}	
	\includegraphics[width=0.06\textwidth]{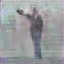}
	\includegraphics[width=0.06\textwidth]{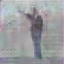} \hspace{0.03cm}
	\includegraphics[width=0.06\textwidth]{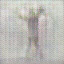}
	\includegraphics[width=0.06\textwidth]{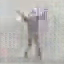}
	\includegraphics[width=0.06\textwidth]{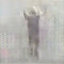}	
	\includegraphics[width=0.06\textwidth]{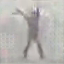}
	\includegraphics[width=0.06\textwidth]{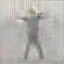} \hspace{0.03cm}
	\includegraphics[width=0.06\textwidth]{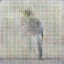}
	\includegraphics[width=0.06\textwidth]{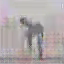}
	\includegraphics[width=0.06\textwidth]{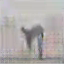}	
	\includegraphics[width=0.06\textwidth]{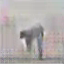}
	\includegraphics[width=0.06\textwidth]{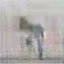}
	\caption{\textbf{Frame samples on Weizmann Action dataset}. Both MoCoGAN (first row) and VOS-GAN (second row) are able to generate realistic videos: the former with higher resolution, but lower motion quality as shown in the quantitative video generation performance.}
	\label{fig:samples_had}
\end{figure*}

On MTurk, each job is created by randomly choosing two of the models under comparison and generating a 16-video batch from each of them; workers are then asked to choose which of the two sets of videos look more realistic. All workers have to provide answers for all the generated batches. We consider only the answers by workers with a lifetime Human Intelligent Task rate over than 90\%. \change{The achieved results are reported in Tab. \ref{tab:qualitative} and show how our method generates more visually-plausible ``golf'' videos compared to VGAN, TGAN, and MoCoGAN. MoCoGAN, instead, outperforms our approach on the Weizmann Action dataset. It should be noted that MoCoGAN, differently from our approach, employs two discriminators --- one for single frames and one for the whole video --- and given the low scene variability in the Weizmann Action dataset, the frame generator is able to produce high-quality frames. % (despite the dataset contains few -90- videos, the number of frames is enough for training a good frame generator). 
In our approach, instead, background and foreground are integrated in videos and the discriminator is mainly trained (see Eq.~\ref{eq:loss_D}) to capture motion features rather than visual appearance. A direct consequence from our learning schema is that VOS-GAN requires many samples to generate good resolution videos, and this explains why performance is better in the ``Golf course'' dataset (about 600,000 videos) than in Weizmann Action dataset (less than 100 videos). However, the capability of our approach to learn better motion cues than MoCoGAN is demonstrated by the results obtained on quantitative generation evaluation (see Sect. \ref{sec:quantitative_gan}) and in action recognition task (see Sect. \ref{sec:vac})}.

\begin{table}[h!]
	\centering
	\begin{tabular}{lc}
		\toprule
		& User preference \% \\
		\midrule
		\multicolumn{2}{c}{\textbf{Golf course}}\\
		\midrule
		VOS-GAN vs VGAN \cite{NIPS2016_6194} & 	 \textbf{80.5} / 19.5\\
		VOS-GAN vs TGAN \cite{Saito_2017_ICCV} & 	\textbf{65.1} / 34.9 \\
		VOS-GAN vs MoCoGAN \cite{Tulyakov_2018_CVPR} & 	\textbf{58.2} / 41.8 \\
		\midrule
		\multicolumn{2}{c}{\textbf{Weizmann Action dataset}}\\
		\midrule
		VOS-GAN vs VGAN \cite{NIPS2016_6194} & 	 \textbf{82.2} / 17.8\\
		VOS-GAN vs TGAN \cite{Saito_2017_ICCV} & 	\textbf{70.3} / 29.7 \\
		VOS-GAN vs MoCoGAN \cite{Tulyakov_2018_CVPR} & 	39.7 / \textbf{60.3} \\
		
		%VOS-GAN vs Real videos & 19.4 / \textbf{80.6}\\
		\bottomrule
	\end{tabular}
	\caption{User preference score (percentage) on video generation quality on different types of generated videos.}\label{tab:qualitative}
\end{table}

Samples of generated videos for VGAN, TGAN, MoCoGAN and our method are shown in Fig. \ref{fig:samples_golf}, while comparisons with MoCoGAN on the Weizmann Action Dataset are shown in Fig.~\ref{fig:samples_had}.

\begin{table*}[h!]
	\centering
	\begin{tabular}{lccccccccc}
		\toprule
		& & & & & & FCD $\downarrow$ & FID $\downarrow$ & IS $\uparrow$\\
		\midrule
		& SLERP 		& LSTM 		& Segmentation & OF & $\alpha$ & 					& 				\\
		\midrule
		Baseline &  	 		&			& 			   &    &          & 5.41	$\pm$ 0.020	& 44.02	$\pm$ 0.45	&  1.98 $\pm$ 0.019\\
		Model 1  &  \checkmark	& 			&			   &    & 		   & 5.27	$\pm$ 0.024	& 41.76	$\pm$ 0.53 	&  2.07 $\pm$ 0.024\\
		Model 2  &	\checkmark  & \checkmark &  			    &		   & & 		4.77	$\pm$ 0.018	& 40.54	$\pm$ 0.41 	&  2.29 $\pm$ 0.019\\%4.61	$\pm$ 0.026	& 38.79	$\pm$ 0.53	&  2.73 $\pm$ 0.011\\
		Model 3  &  \checkmark  & \checkmark & \checkmark  &    &		   & 4.61	$\pm$ 0.026	& 38.79	$\pm$ 0.53	&  2.73 $\pm$ 0.011\\%4.39 $\pm$ 0.014	& 37.48	$\pm$ 0.39	&  2.88 $\pm$ 0.021\\
		Model 4  &  \checkmark  & \checkmark & \checkmark  & \checkmark &  & 4.22	$\pm$ 0.022 & 33.18 $\pm$ 0.43	& 3.09 $\pm$ 0.011\\
		VOS-GAN  &  \checkmark  & \checkmark & \checkmark  & \checkmark & \checkmark & \textbf{4.11 $\pm$ 0.018}	&\textbf{31.32 $\pm$ 0.46}    & \textbf{3.16 $\pm$ 0.032}\\
		%		VOS-GAN (w/o LSTM)		& 4.61	$\pm$ 0.026	& 38.79	$\pm$ 0.53	&  2.73 $\pm$ 0.011\\%
		%		VOS-GAN (w/o segm.)		& 4.39 $\pm$ 0.014	& 37.48	$\pm$ 0.39	&  2.88 $\pm$ 0.021\\
		\bottomrule
		
	\end{tabular}
	\caption{\textbf{Ablation studies.} Quantitative evaluation on the ``Golf course'' dataset of different configurations of the proposed model. The baseline is the GAN architecture described in Sect. \ref{sec:model} trained with traditional adversarial loss and using a random latent variable for foreground content in addition to the bakcground latent space.}\label{tab:quantitative_golf}
\end{table*}

\begin{table}[h!]
	\centering
	\begin{tabular}{lccc}
		\toprule
		& FCD $\downarrow$ & FID $\downarrow$ & IS $\uparrow$\\
		\midrule
		\multicolumn{4}{c}{\textbf{Golf course}}\\
		\midrule
		VGAN 	& 10.61 $\pm$ 0.015	& 45.32 $\pm$ 0.31	& 1.74 $\pm$ 0.021\\
		TGAN    & 5.74  $\pm$ 0.025	& 42.58 $\pm$ 0.39	& 2.02 $\pm$ 0.019\\
		MoCoGAN & 5.01   $\pm$	0.023& 34.53 $\pm$ 0.42	& 2.47 $\pm$ 0.013\\
		VOS-GAN				& \textbf{4.11 $\pm$ 0.018}	&\textbf{31.32 $\pm$ 0.46}    & \textbf{3.16 $\pm$ 0.032}\\
		\midrule	
        	
		\multicolumn{4}{c}{\textbf{Weizmann Action dataset}}\\
		\midrule
		VGAN 	& 5.18  $\pm$ 0.021		& 7.64 $\pm$ 0.041	& 2.78 $\pm$ 0.027\\
		TGAN    & 4.53  $\pm$ 0.029		& 7.00 $\pm$ 0.027	& 2.94 $\pm$ 0.016\\
		MoCoGAN & \textbf{4.07  $\pm$ 0.018} 	& 5.74 $\pm$ 0.031    &  \textbf{3.76 $\pm$ 0.025}\\
		VOS-GAN	& 4.24  $\pm$ 0.016		& \textbf{5.71 $\pm$ 0.034} 	& 3.29 $\pm$ 0.020	\\
\bottomrule

	\end{tabular}
	\caption{Comparison of quantitative generation performance, in terms of foreground content distance (FCD), Fr\'echet Inception Distance (FID) and Inception Score (IS), against VGAN, TGAN and MoCoGAN, respectively, on the ``Golf course'' and the Weizmann action dataset.}\label{tab:quantitative_comparison}
\end{table} 

\begin{table*}[!htbp]
	\centering
	\begin{tabular}{llcccccccccc}
	\toprule
	\textbf{Learning} & \textbf{Model} &
	&\multicolumn{2}{c}{\textbf{DAVIS-16}}& 
	&\multicolumn{2}{c}{\textbf{DAVIS-17}}& & \multicolumn{2}{c}{\textbf{SegTrack-v2}} & \\%& \multicolumn{2}{c}{\textbf{F4K-Fish}}\\
	& & &$\mathcal{M}_{\mathcal{F}}$ & $\mathcal{M}_{\mathcal{J}}$ & &$\mathcal{M}_{\mathcal{F}}$ & $\mathcal{M}_{\mathcal{J}}$ &&  $\mathcal{M}_{\mathcal{F}}$ & $\mathcal{M}_{\mathcal{J}}$ \\%&%&  $\mathcal{M}_{\mathcal{F}}$ \\%& $\mathcal{M}_{\mathcal{J}}$ \\
	\midrule
	Unsupervised & Synthetic VOS && 	25.41 & 33.66 & & 19.83 & 21.87 & &20.32 & 23.72 & \\%&20.63 & 23.13\\
    & Adversarial VOS &&	31.22 & 38.11 & & 22.57 & 27.01 & & 24.11&27.42 & \\%&  22.12 & 25.59\\
    \midrule
	Supervised & Synthetic VOS FT &&	60.85 & 64.66 & & 52.54  & 55.12 && 56.43 & 59.95 &\\%& 49.54 & 55.07\\
    & Adversarial VOS FT & & \textbf{67.35} & \textbf{71.24} & & \textbf{56.10} & \textbf{61.65} && \textbf{61.14} & \textbf{65.02} &\\%& \textbf{56.31} & \textbf{61.12}\\
    \midrule
    & Baseline  &&	    57.05 & 62.85 & & 50.96 & 53.41  && 53.26 & 57.03 &\\%& 48.64 & 53.81\\
\bottomrule
	\end{tabular}
\caption{Video object segmentation results (in percentage). The first two rows report the results obtained by training the model without annotations, while the third and forth rows report the performance when fine-tuning on the video benchmarks. The last row shows the results achieved by our baseline trained purely in a supervised way.}\label{tab:vos_results}
\end{table*}

\subsubsection{Quantitative evaluation}
\label{sec:quantitative_gan}
\change{Quantitative evaluation of video generation performance is carried out by measuring FCD, FID and IS scores on 20 sets of 50,000 videos generated by the compared models trained on ``Golf course'' of~\cite{NIPS2016_6194}, and on 20 sets of 500 videos generated on the Weizmann Action Dataset. Results are computed in terms of mean and standard deviation of each metrics over the sets of generated samples.\\
Firstly, we perform an ablation study on ``Golf course'' to understand how our GAN design choices affect the quality of the generated videos, by evaluating the above-mentioned metrics when each proposed term is included in the model. In particular, we computed the performance of VOS-GAN excluding all components (i.e., loss on segmentation and optical flow and replacing the RNN-SLERP based trajectory latent space modeling with a random latent space) --- VOS-GAN (baseline) --- and when gradually including individual components.
The results of our ablation study are given in Tab. \ref{tab:quantitative_golf}. Both loss terms (on segmentation and optical flow) contribute to the model's accuracy: optical flow has the largest impact, likely because foreground regions usually correspond to clusters of oriented vectors in the optical flow, hence learning to compute it accurately is also informative from the segmentation perspective. Both SLERP and LSTM also contribute significantly to generating visually-plausible videos by modeling better motion as demonstrated by the increase in FID and FCD metrics.\\
Tab.~\ref{tab:quantitative_comparison} shows the comparison with existing methods on the ``Golf course'' and on the Weizmann Action datasets. The results show that VOS-GAN outperforms VGAN, TGAN and MoCOGAN on the three metrics on ``Golf course'', while MoCoGAN is better than the proposed approach, on two out of the three adopted metrics, on the Weizmann Action Dataset.
The reasons behind the different behavior of VOS-GAN and MoCoGAN on the two datasets are similar to those mentioned above in the qualitative analysis: 1) The Weizmann action dataset is characterized by few videos with constrained motion and background, thus the variability among frames is low and MoCoGAN's image generator is able to model scene appearance with good quality. On the contrary, our approach is able to learn motion (as demonstrated by the results in action recognition given below), but it does not receive enough training data (given the size of the dataset) for learning well the scene; 2) ``Golf course'' contains many video sequences with high variability in terms of motion and appearance --- variability that MoCoGAN is not able to learn. However, VOS-GAN confirms its capabilities to learn better motion, indeed, it achieves a higher FID than MoCoGAN, substantiating also our previous claims on the different performance of MoCoGAN on the two employed datasets.} 

%\begin{table}[h!]
%	\centering
%	\begin{tabular}{lccc}
%		\toprule
%		& FCD & MC & IS\\
%		\midrule
%		MoCoGAN \cite{Tulyakov_2018_CVPR} 	 & 4.15		& 0.009		& \textbf{3.73}\\
%		VOS-GAN\textsubscript{W} (w/o OF)	 & 4.83		& 0.019		& 2.97\\
%		VOS-GAN\textsubscript{W} (w/o segm.) & 4.41		& 0.011		& 3.05\\
%		VOS-GAN\textsubscript{W}			& \textbf{3.88}	&\textbf{0.006}	& 3.26\\
%		%Real videos 			            & 3.76		& 0.0001	& 4.91\\
%%%	\caption{Quantitative evaluation on the Weizmann Action dataset. Metrics and models as described in Tab.~\ref{tab:quantitative_golf}.}\label{tab:quantitative_had}
%\end{table}

\subsection{Video Object Segmentation}
The first part of this evaluation aims at investigating the contribution of adversarial training for video object segmentation, by assessing the quality of the foreground maps computed by our video object segmentation approach in four different settings:
\begin{itemize}
\item \emph{synthetic}: we use the motion stream, i.e. segmentation subnetwork, from the architecture of our discriminator and train it from scratch with the foreground masks synthesized by our generator trained on ``Golf course'' video dataset; 
\item \emph{adversarial}: we use segmentation subnetwork
from the discriminator of VOS-GAN trained on the ``Golf dataset'';
\item \emph{fine-tuned synthetic}: the segmentation network trained in the \emph{synthetic} modality is then fine-tuned on ground-truth masks of the benchmark datasets' (DAVIS 2016/2017 and SegTrack-v2) training sets; 

\item \emph{fine-tuned adversarial}: analogously, we use the segmentation model from the \emph{adversarial} training scenario and fine-tune it on real segmentation masks. 
\end{itemize}

As baseline we select the segmentation network of our GAN model trained, in a supervised way (i.e., using annotations at training time), from scratch on the training splits of the employed datasets. As metrics, we use mean values for region similarity and contour accuracy, i.e.,  $\mathcal{M}_{\mathcal{J}}$ and $\mathcal{M}_{\mathcal{F}}$.

\begin{figure*}[!htbp]
 	\centering
	\includegraphics[width=0.15\textwidth]{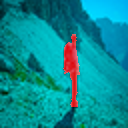}
	\includegraphics[width=0.15\textwidth]{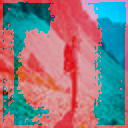}
	\includegraphics[width=0.15\textwidth]{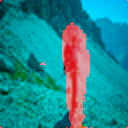} \hspace{0.2cm}
	\includegraphics[width=0.15\textwidth]{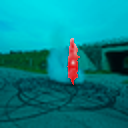}
	\includegraphics[width=0.15\textwidth]{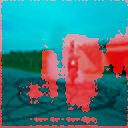}
	\includegraphics[width=0.15\textwidth]{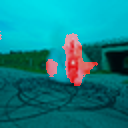}\\ \vspace{0.2cm}
	\includegraphics[width=0.15\textwidth]{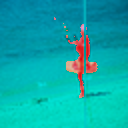}
	\includegraphics[width=0.15\textwidth]{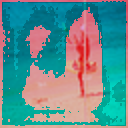}
	\includegraphics[width=0.15\textwidth]{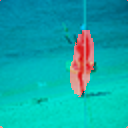} \hspace{0.2cm}
	\includegraphics[width=0.15\textwidth]{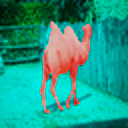}
	\includegraphics[width=0.15\textwidth]{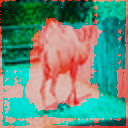}
	\includegraphics[width=0.15\textwidth]{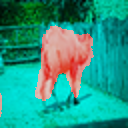}\\ \vspace{0.6cm}
	\includegraphics[width=0.15\textwidth]{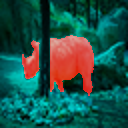}
	\includegraphics[width=0.15\textwidth]{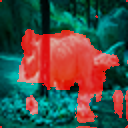}
	\includegraphics[width=0.15\textwidth]{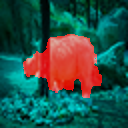} \hspace{0.2cm}
	\includegraphics[width=0.15\textwidth]{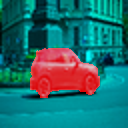}
	\includegraphics[width=0.15\textwidth]{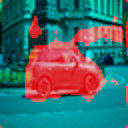}
	\includegraphics[width=0.15\textwidth]{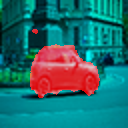}\\ \vspace{0.2cm}
	\includegraphics[width=0.15\textwidth]{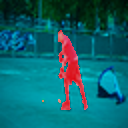}
	\includegraphics[width=0.15\textwidth]{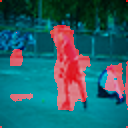}
	\includegraphics[width=0.15\textwidth]{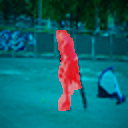} \hspace{0.2cm}
	\includegraphics[width=0.15\textwidth]{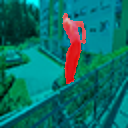}
	\includegraphics[width=0.15\textwidth]{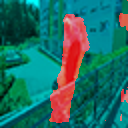}
	\includegraphics[width=0.15\textwidth]{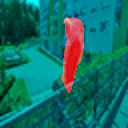}\\ \vspace{0.6cm}
	\includegraphics[width=0.15\textwidth]{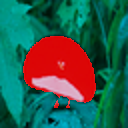}
	\includegraphics[width=0.15\textwidth]{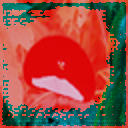}
	\includegraphics[width=0.15\textwidth]{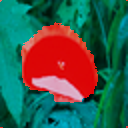} \hspace{0.2cm}
	\includegraphics[width=0.15\textwidth]{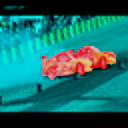}
	\includegraphics[width=0.15\textwidth]{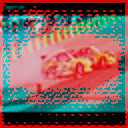}
	\includegraphics[width=0.15\textwidth]{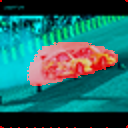}\\ \vspace{0.2cm}
	\includegraphics[width=0.15\textwidth]{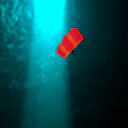}
	\includegraphics[width=0.15\textwidth]{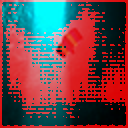}
	\includegraphics[width=0.15\textwidth]{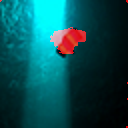} \hspace{0.2cm}
	\includegraphics[width=0.15\textwidth]{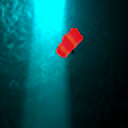}
	\includegraphics[width=0.15\textwidth]{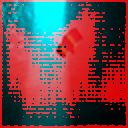}
	\includegraphics[width=0.15\textwidth]{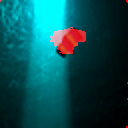}\\ \vspace{0.4cm}
 	\caption{\textbf{Video object segmentation results on multiple datasets}. First two rows: DAVIS 2017; second two rows: DAVIS 2016; last two rows: SegTrack-v2. Each image block shows the original image with ground truth, the image with the map obtained by the adversarially trained VOS-GAN and the image with the segmentation mask when fine-tuning the model on the given dataset. Images are shown at the original resolution provided by our models, i.e., 64 $\times$ 64. The best segmentation masks are obtained on DAVIS 2016, as also demonstrated by the results in Table \ref{tab:vos_results}. Our purely unsupervised approach is still far from the performance of supervised training ones, but it is able to detect object movements in videos characterized by strong camera motion (as those hereby reported).
 		%It can be noted that the adversarial trained VOS detected all (both background and foreground) motion, while fine-tuning had the effect to focus only on objects of interest. 
 	}
 	\label{fig:vos}
 \end{figure*}

\begin{figure*}[h!]
 	\centering
 	\includegraphics[width=0.41\textwidth]{images/davis2017_j.pdf}  	
	\includegraphics[width=0.41\textwidth]{images/davis2017_f.pdf}\\%\\ \vspace{0.2cm}
	\includegraphics[width=0.41\textwidth]{images/davis2016_j.pdf} 
 	\includegraphics[width=0.41\textwidth]{images/davis2016_f.pdf}\\%\\ \vspace{0.2cm}
 	\includegraphics[width=0.41\textwidth]{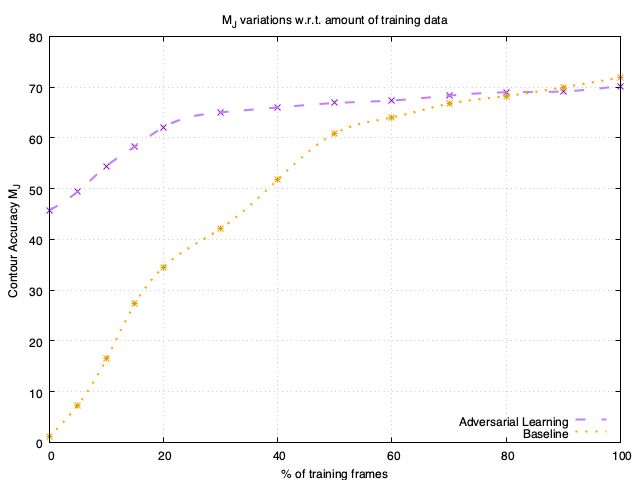} 	 	
	\includegraphics[width=0.41\textwidth]{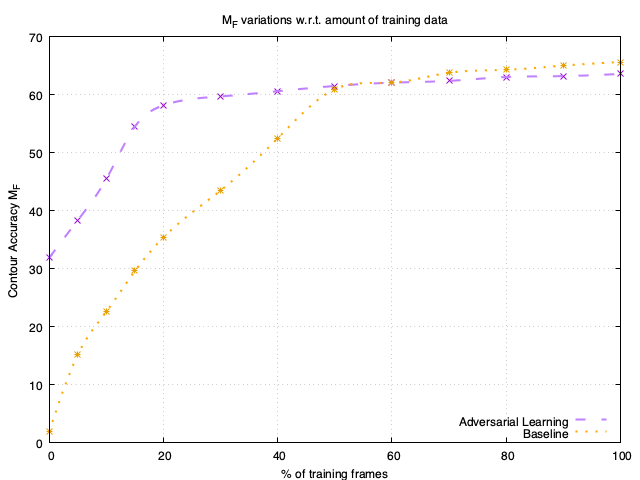}  	
 	 	\caption{Performance, in terms of region similarity $\mathcal{M}_\mathcal{J}$ and contour accuracy $\mathcal{M}_\mathcal{F}$, w.r.t. to the percentage of training images of DAVIS 2017 (first row), DAVIS 2016 (second row) and SegTrack-v2 (third row) datasets. Note that ``\% of training frames'' is related to the size of each video benchmark employed in the evaluation.
 	 	%indicates the percentage of training frames - sampled uniformly from all training videos- of the employed video benchmarks. 
 		%It can be noted that the adversarial trained VOS detected all (both background and foreground) motion, while fine-tuning had the effect to focus only on objects of interest. 
 	}
 	\label{fig:performance_training_size}
 \end{figure*}

Tab. \ref{tab:vos_results} shows the obtained performance: our segmentation network obtains fair results even when trained without labeled data (first two rows of the table). There is a significant gap between region similarity $\mathcal{M}_\mathcal{J}$ (higher) and contour accuracy $\mathcal{M}_\mathcal{F}$ (lower), meaning that the model performs better in motion localization than in object segmentation. \change{Results also show that our adversarial model fine-tuned on real data (fourth line) significantly outperforms, by a margin of about 10\%, our baseline (fifth line), indicating that pre-training, with self-supervision, the segmentation model obtains better accuracy. 
Pre-training the segmentation model with synthetic videos before fine-tuning on real data (third line) leads also to increased (about 2\%) performance than the baseline. Thus, our video generation model acts as a data distillation approach \cite{RadosavovicDGGH18}, i.e., through injecting synthetic video masks into the training procedure, our approach combines predictions from arbitrary transformations (enabled by the generator) of unlabeled videos.
We further quantify the percentage of training examples needed to obtain satisfactory results both for our baseline and the adversarial fine-tuned model. The results on the three employed datasets are given in Fig. \ref{fig:performance_training_size} and indicate that our adversarial model requires fewer annotations to reach satisfactory performance than our baseline, which requires at least twice as many annotations.
The obtained results, thus, demonstrate that our approach not only yields better segmentation results, but it also allows us to reduce the needed training data.
Fine-tuning the segmentation subnetwork on real data (third and fourth rows in Tab. \ref{tab:vos_results})  has the effect to focus better on target objects and to cope with camera motion (that is not considered by the video generation model), as demonstrated by the performance improvement reported in rows three and four. This effect is also shown in Fig.~\ref{fig:vos}, that shows that the segmentation maps predicted by our purely-unsupervised method highlight object motion, but object contours are not accurate. 
Training the segmentation network in an adversarial framework (rather than directly on the fake maps) leads to a  significant performance increase, both with and without fine-tuning on real data. Fine-tuning the whole GAN video generation on the three employed video benchmarks, instead, does not yield a significant performance improvement, but relaxes the requirements on available manual annotations.}

 \begin{figure*}[h!]
 	\centering
 	\includegraphics[width=0.85\textwidth]{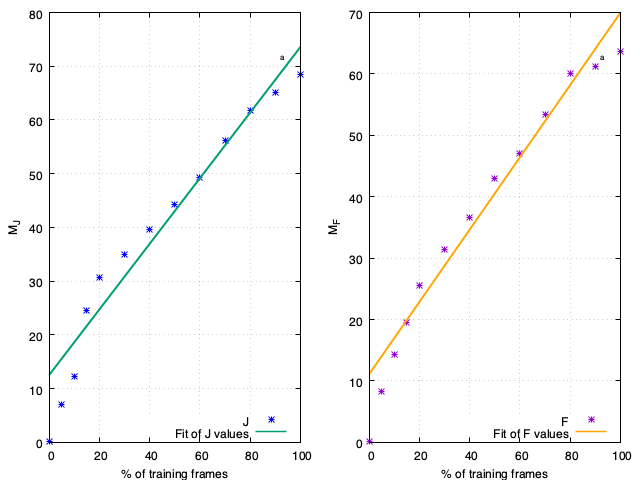}  		
 	\caption{Region similarity $\mathcal{M}_\mathcal{J}$ and contour similarity $\mathcal{M}_\mathcal{F}$ w.r.t. to the percentage of videos used for training our GAN-based video generation model fine-tuned on the DAVIS 2017 dataset. Line fitting is performed by least square regression.}
 	\label{fig:video_gen_training_size}
 \end{figure*}

On the other hand, the advantage of using less annotated data comes at the cost of training the GAN for video generation using unlabeled videos. For this reason, we quantify how the amount of unlabeled videos for video generation affects video object segmentation performance. This evaluation is carried out on the adversarial configuration for video object segmentation, and uses region similarity and contour accuracy as metrics. Fig. \ref{fig:video_gen_training_size} shows an almost linear dependence between video object segmentation accuracy and the number of unlabeled videos used for generation by our VOS-GAN, suggesting that using more unlabeled videos in training the adversarial models may ideally lead to better results as the approach effectively learns motion dynamics.

\change{Since our proposed adversarial framework is used for unsupervised pre-training video segmentation models, we compare it, on DAVIS 2017, to: a) two approaches that perform optical flow estimation (both non-learning-based \cite{Farneback:2003} and learning-based \cite{IMSKDB17} optical flow) and assign foreground pixels based on the empirically-optimal threshold over flow magnitude; b) the approach proposed in \cite{VondrickSFGM18}, relaxing its constraint of having the first annotated label at inference time by replacing the needed frame with a binary image obtained by thresholding the optical flow between the first two frames, as done in the previous methods. 
We also compare the performance of our adversarial learning method fine-tuned on DAVIS 2017 to our discriminator network trained adversarially to estimate only optical flow and then fine-tuned on the video segmentation task ("OF estimation + FT" in Table \ref{tab:un_vos1}). Results, in terms of $M_{\mathcal{J}}$ and  $M_{\mathcal{F}}$, are given in Table \ref{tab:un_vos1}.
Our purely adversarial unsupervised approach learns better features (segmentation performance higher by about 8-10\%) than both optical flow methods and \cite{VondrickSFGM18}. %The peformance are better in $J$ metric than in $F$ as our generator performs better in modeling motion than object appearance.
Also, when fine-tuned, our approach gives better segmentation performance than our generation model trained without segmentation mask self-supervision (i.e., the GAN model using only optical flow) suggesting that, indeed, our self-supervision strategy through synthetic masks during video generation learns better representations for the segmentation task.}

\begin{table}[h!]
	\centering
	\begin{tabular}{cccc}
		\toprule
		& Method & $M_{\mathcal{J}}$ & $M_{\mathcal{F}}$\\
		\midrule
		Unsupervised & Optical Flow \cite{Farneback:2003} & 14.55 & 10.32\\
		& Optical Flow \cite{IMSKDB17} & 17.21 & 15.43\\
		& Unsupervised \cite{VondrickSFGM18} & 19.86& 17.47\\
		& Adversarial VOS & \textbf{27.01} & \textbf{22.57}\\
		\midrule
		Supervised & OF estimation + FT & 54.32 & 47.73\\
		& Adversarial VOS + FT & \textbf{61.65} & \textbf{56.10}\\		
		\bottomrule
	\end{tabular}
	\caption{Performance on DAVIS 2017 by unsupervised (i.e., not using any annotated data at inference time) video object segmentation methods, as well as of unsupervised models fine-tuned on DAVIS 2017.}\label{tab:un_vos1}
\end{table}

\change{We finally compare our approach to state-of-the-art unsupervised video object segmentation methods (i.e., not using any annotations at test time), namely, 	\cite{BMVC.28.21}, \cite{papazoglou2013fast}, \cite{Tokmakov17}, \cite{JainXG17}, \cite{8100267} on DAVIS 2017; \cite{BMVC.28.21}, \cite{papazoglou2013fast},  \cite{Jang2018VideoPW} and \cite{8100267} on DAVIS 2016; \cite{papazoglou2013fast},  \cite{6248065}, \cite{BMVC2015_186}, \cite{HallerL17} on SegTrack-v2. For a fair comparison we did not test any of the existing code, but just report performance from the literature, specifically from \cite{Man+18b} for DAVIS 2016, from \cite{8100267} for DAVIS 2017 and from \cite{HallerL17} for SegTrack-v2.
The achieved results are given in Tab. \ref{tab:vos_comparison} and show that our approach is almost on par with the best performing method on the DAVIS benchmarks and outperforms existing methods on SegTrack-v2. Most importantly, according to Fig. \ref{fig:performance_training_size} and results in Table \ref{tab:vos_comparison}, it leads to performance enhancement requiring less annotated data.} %Thus, 
%combining any segmentation model with the proposed framework will result in a significant segmentation performance increase.}
%our framework defines a promising paradigm for improving the accuracy of any segmentation approach.}
%We also compare our results on Segtrack-v2 to those of state of the art methods, i.e., \cite{papazoglou2013fast},  \cite{6248065}, \cite{BMVC2015_186}, \cite{HallerL17}, obtaining higher performance as shown in Table \ref{tab:segv2}.}%, indicate that our method outperforms approaches on the SegTrack-v2 dataset, 
%while it performs worse on the F4K benchmark. This can be explained with fundamental difference between motion of common objects (as found in the DAVIS16/17 and SegTrack-v2) and fish motion. In addition objects in F4K are tiny, thus, when downsampling to $64x64$ frame size (to make it work with our model) it becomes very complex to be accurate.}%\footnote{Testing this claim on existing models is out of the scope of the paper. Furthemore, the comparing approaches does either release code or code is not written in Pytorch as ours.}.}

\begin{table}[h!]
	\centering
	\begin{tabular}{lcccccc}
		\toprule
		\multicolumn{7}{c}{\textbf{DAVIS 2016}}\\
        \midrule
		&\cite{BMVC.28.21} & \cite{papazoglou2013fast} & \cite{Tokmakov17} & \cite{JainXG17} & \cite{8100267} & Ours\\
		\midrule
        \multicolumn{7}{l}{\textbf{$\mathcal{J}$ - Region Similarity}}\\
		\midrule
        Mean $\mathcal{M}$   &64.1   &57.5   & 70.0 & 70.7   & \textbf{76.3}   & \textit{71.2}\\
		Recall $\mathcal{O}$ &73.1   &65.2   & \textit{85.0} & 83.5   & \textbf{89.2}   & 84.6\\
		%Decay $\mathcal{R}$  &2.4    &0.0    &12.6    &1.3   &1.5  & 7.0&2.1\\
		\midrule
        \multicolumn{7}{l}{\textbf{$\mathcal{F}$ - Contour accuracy}}\\
		\midrule
        Mean $\mathcal{M}$   &59.3   &53.6   & 65.9 & 65.3   & \textbf{71.1}  & \textit{67.3}\\
		Recall $\mathcal{O}$ &65.8   &57.9   & \textit{79.2} & 73.8   & \textbf{82.8}  & 77.9\\
		%Decay $\mathcal{R}$  &5.1    &2.9    &11.4    &2.5    &1.8  & 7.9&1.4\\
		\midrule
		\multicolumn{7}{l}{\textbf{$\mathcal{J} \& \mathcal{F}$ -  \textbf{Average}}}\\
		\midrule
        Mean $\mathcal{M}$   &61.7   &55.5   & 67.9 & 68.0   & \textbf{73.7}  & \textit{69.2}\\
		Recall $\mathcal{O}$ &69.4   &61.5   & \textit{82.1} & 78.6   & \textbf{86.0}  & 81.2\\
		\bottomrule
	    \multicolumn{7}{c}{\textbf{}}\\
	    \toprule
		\multicolumn{7}{c}{\textbf{DAVIS 2017}}\\
        \midrule
		& & \cite{BMVC.28.21} & \cite{papazoglou2013fast} & \cite{Jang2018VideoPW} & \cite{8100267}  &Ours\\
		\midrule
        \multicolumn{7}{l}{\textbf{$\mathcal{J}$ - Region Similarity}}\\
		\midrule
        Mean $\mathcal{M}$   &&51.4   &49.6      & 45.0 &\textbf{63.3 } & \textit{61.7}\\
		Recall $\mathcal{O}$ &&55.5   &52.9      & 46.4 &\textbf{72.9} &\textit{61.4}\\
		%Decay $\mathcal{R}$  &2.4    &0.0    &12.6    &1.3   &1.5  & 7.0&2.1\\
		\midrule
        \multicolumn{7}{l}{\textbf{$\mathcal{F}$ - Contour accuracy}}\\
		\midrule
        Mean $\mathcal{M}$   &&48.6   &48.0  &44.8  & \textbf{61.2} & \textit{56.1}\\
		Recall $\mathcal{O}$ &&49.4   &46.8  &43.0  & \textbf{67.8} & \textit{53.8}\\
		\midrule
		\multicolumn{7}{l}{\textbf{$\mathcal{J} \& \mathcal{F}$ - Average}}\\
		\midrule
        Mean $\mathcal{M}$   && 50.0 & 48.8 & 44.9   &  \textbf{62.2}  & \textit{58.9}\\
		Recall $\mathcal{O}$ &&  52.4& 49.8 & 44.7   &  \textbf{70.3}  & \textit{57.6}\\
		\midrule
		\multicolumn{7}{c}{\textbf{SegTrack-v2}}\\
		\midrule
		& &\cite{papazoglou2013fast} & \cite{6248065} & \cite{BMVC2015_186} & \cite{HallerL17} & Ours\\
		\midrule
		\multicolumn{7}{l}{\textbf{$\mathcal{J}$ - Region Similarity}}\\
		\midrule
		Mean $\mathcal{M}$   &&50.1 & 30.4 & 49.9 & \textit{61.1} & \textbf{65.0}\\
		\bottomrule
	\end{tabular}
	\caption{Comparison to state-of-the-art unsupervised methods on DAVIS 2016, DAVIS 2017 and SegTrack-v2 benchmarks in terms of  region similarity ($\mathcal{J}$) contour accuracy ($\mathcal{F}$). On SegTrack-v2 we report only mean $\mathcal{J}$ as comparing methods employ only that measure since recall  and contour accuracy metrics were introduced only in \cite{7780454}.
	In bold best performance, in italic the second best performance. }\label{tab:vos_comparison}
\end{table}

%\begin{table}[h!]
%	\centering
%	\begin{tabular}{ccccc}
%		\toprule
%		\multicolumn{5}{c}{\textbf{Segtrack-v2}}\\
%		\midrule
%		\cite{papazoglou2013fast} & \cite{6248065} & \cite{BMVC2015_186} & \cite{HallerL17} & Ours\\
%		\midrule
%		50.1 & 30.4 & 49.9 & \textit{61.1} & \textbf{65.0}\\
%		\bottomrule	
%\end{tabular}
%	\caption{Comparison, in terms of  region similarity $\mathcal{J}$, to state-of-the-art unsupervised methods (performance taken from \cite{HallerL17}) on the Segtrack-v2 and F4K benchmarks. In bold best performance, in italic the second best performance.}\label{tab:segv2}
%\end{table}

%\begin{table}[h!]
%	\centering
%	\begin{tabular}{ccccc}
%		\toprule
%		\multicolumn{5}{c}{\textbf{F4K}}\\
%		\midrule
%		\cite{papazoglou2013fast} & \cite{5672785} & \cite{SPAMPINATO201474} & \cite{7299114} &Ours\\
%		\midrule
%		34.3 & 76.7 & \textit{81.8} & \textbf{86.9} & 70.9\\
%%		\bottomrule	
%	\end{tabular}
%	\caption{Comparison, in terms of $F_1$ measure, to state-of-the-art unsupervised methods (values taken from \cite{7299114}) on the F4K benchmark.	In bold best performance, in italic the second best performance.}\label{tab:f4k}
%\end{table}

\subsection{Video Action Recognition}\label{sec:vac}
Finally, in order to evaluate the representational power of features learned by the methods under comparison, we employ the models' discriminators (after the initial training for video generation) to perform action recognition on the UCF101 and Weizmann Action datasets.  Furthermore, results on video action recognition serve also as an additional means to verify video generation performance especially in learning motion dynamics. 

This analysis is carried out on two different training settings: a)~\emph{transfer learning}: each model's discriminator is used as a feature extractor by removing the final real/fake output layer and training a linear classifier on the exposed features; b)~\emph{fine-tuning}: the real/fake output layer is replaced with a softmax layer and the whole discriminator is fine-tuned for classification. %The action recognition models are trained for 40 epochs on UCF101 and for 15 epochs on Weizmann Action dataset, using ADAM as optimizer.

In order to make the comparison fair and consistent with the initial adversarial training, we separately evaluate the models originally trained on ``Golf course'' from those trained on Weizmann Action dataset: hence, Tab.~\ref{tab:var_results_golf_to_ucf} and \ref{tab:var_results_golf_to_weizmann} report the classification accuracy on UCF101 and Weizmann Action dataset by comparing the discriminators originally trained for video generation on ``Golf course''; similarly, Tab.~\ref{tab:var_results_weizmann_to_ucf} and \ref{tab:var_results_weizmann} report the classification accuracy by the discriminators originally trained on Weizmann Action dataset. The results show that our VOS-GAN discriminator outperforms VGAN, TGAN and MoCoGAN in all the considered scenarios.
%In Tab.~\ref{tab:var_results_golf_to_ucf} and \ref{tab:var_results_golf_to_weizmann}, lower accuracy scores compared to the ones reported in \cite{NIPS2016_6194} are explained by the fact that we train the models only on ``Golf course'' videos (600,000 sequences), while \cite{NIPS2016_6194} is trained on many more videos (over 35 million). 
Note that all models quickly overfit on Weizmann Action dataset, due to the small size of the dataset, which likely explains why transfer learning performs better than fine-tuning.

\begin{table}[h!]
 \centering
	\begin{tabular}{lccc}
		\toprule
		\textbf{Settings} &\multicolumn{3}{c}{\textbf{Models}}\\
		\midrule
		& VGAN & TGAN & VOS-GAN\textsubscript{G} \\
		\midrule
		Transfer learning & 39.19 & 32.45 & \textbf{41.02} \\ 
		Fine-tuning       & 45.43 & 36.94 & \textbf{49.33} \\
		\bottomrule
	\end{tabular}
 \caption{Video action recognition accuracy on UCF101 of the models originally trained on ``Golf course'' (classification accuracy, in percentage).}
 \label{tab:var_results_golf_to_ucf}
\end{table}

\begin{table}[h!]
 \centering
	\begin{tabular}{lcc}
		\toprule
		\textbf{Settings} &\multicolumn{2}{c}{\textbf{Models}}\\
		\midrule
		& MoCoGAN & VOS-GAN\textsubscript{W} \\
		\midrule
		Transfer learning & 18.47 & \textbf{29.42}\\ 
		Fine-tuning & 32.83 & \textbf{45.66}\\
		\bottomrule
	\end{tabular}
 \caption{Video action recognition accuracy on UCF101 of the models originally trained on Weizmann Action dataset (classification accuracy, in percentage).}
 \label{tab:var_results_golf_to_weizmann}
\end{table}

\begin{table}[h!]
 \centering
	\begin{tabular}{lccc}
		\toprule
		\textbf{Settings} &\multicolumn{3}{c}{\textbf{Models}}\\
		\midrule
		& VGAN & TGAN & VOS-GAN\textsubscript{G} \\
		\midrule
		Transfer learning & 65.87 & 54.19 & \textbf{68.79}\\ 
		Fine-tuning       & 64.41 & 52.30 & \textbf{66.80}\\
		\bottomrule
	\end{tabular}
 \caption{Video action recognition accuracy on Weizmann Action dataset of the models originally trained on ``Golf course'' (classification accuracy, in percentage).}
 \label{tab:var_results_weizmann_to_ucf}
\end{table}

\begin{table}[h!]
 \centering
	\begin{tabular}{lcc}
		\toprule
		\textbf{Settings} &\multicolumn{2}{c}{\textbf{Models}}\\
		\midrule
		& MoCoGAN & VOS-GAN\textsubscript{W} \\
		\midrule
		Transfer learning & 70.76 & \textbf{74.29}\\ 
		Fine-tuning & 67.02 & \textbf{71.63}\\
		\bottomrule
	\end{tabular}
 \caption{Video action recognition accuracy on Weizmann Action Dataset of the models originally trained on the same dataset (classification accuracy, in percentage).}
 \label{tab:var_results_weizmann}
\end{table}

\label{sec:performance}

\section{Conclusion}
We propose a framework for unsupervised learning of motion cues in videos. It is based on an adversarial video generation approach --- VOS-GAN --- that disentangles background and foreground dynamics. Motion generation is improved by learning a suitable object motion latent space and by  controlling the discrimination process with pixelwise dense prediction of moving objects through a self-supervision mechanism, in which segmentation masks are internally synthesized by the generator. 

Extensive experimental evaluation showed that our VOS-GAN outperforms existing video generation methods, namely, VGAN~\cite{NIPS2016_6194}, TGAN~\cite{Saito_2017_ICCV}, MoCoGAN~\cite{Tulyakov_2018_CVPR}, especially in modeling object motion. 
The capability of our approach to better model object motion is further demonstrated by the fact that the learned (in an unsupervised way) features can be used for effective video object segmentation and video action recognition tasks.

Finally, it has to be noted that, although we introduce a new segmentation network as part of the adversarial framework, it is quite general and can easily integrated into any segmentation model, by adding the optical flow dense prediction stream and using masks synthesized by the generator for supervision.

\bibliographystyle{splncs04}
\bibliography{egbib}

\begin{thebibliography}{10}
\providecommand{\url}[1]{\texttt{#1}}
\providecommand{\urlprefix}{URL }
\providecommand{\doi}[1]{https://doi.org/#1}

\bibitem{pmlr-v70-arjovsky17a}
Arjovsky, M., Chintala, S., Bottou, L.: {W}asserstein generative adversarial
  networks. In: ICML (2017)

\bibitem{Bousmalis_2017_CVPR}
Bousmalis, K., Silberman, N., Dohan, D., Erhan, D., Krishnan, D.: Unsupervised
  pixel-level domain adaptation with generative adversarial networks. In: CVPR
  (2017)

\bibitem{Brox:2010}
Brox, T., Malik, J.: Object segmentation by long term analysis of point
  trajectories. In: ECCV (2010)

\bibitem{Cae17}
Caelles, S., Maninis, K.K., Pont-Tuset, J., Leal-Taix\'e, L., Cremers, D., {Van
  Gool}, L.: One-shot video object segmentation. In: CVPR (2017)

\bibitem{NIPS2015_5773}
Denton, E.L., Chintala, S., szlam, a., Fergus, R.: Deep generative image models
  using a laplacian pyramid of adversarial networks. In: Cortes, C., Lawrence,
  N.D., Lee, D.D., Sugiyama, M., Garnett, R. (eds.) NIPS (2015)

\bibitem{doersch2015unsupervised}
Doersch, C., Gupta, A., Efros, A.A.: Unsupervised visual representation
  learning by context prediction (2015)

\bibitem{BMVC.28.21}
Faktor, A., Irani, M.: Video segmentation by non-local consensus voting. In:
  BMVC (2014)

\bibitem{Farneback:2003}
Farneb\"{a}ck, G.: Two-frame motion estimation based on polynomial expansion.
  In: Proceedings of the 13th Scandinavian Conference on Image Analysis. pp.
  363--370. SCIA'03, Springer-Verlag, Berlin, Heidelberg (2003)

\bibitem{6247883}
Fragkiadaki, K., Zhang, G., Shi, J.: Video segmentation by tracing
  discontinuities in a trajectory embedding. In: CVPR (2012)

\bibitem{7299114}
{Giordano}, D., {Murabito}, F., {Palazzo}, S., {Spampinato}, C.:
  Superpixel-based video object segmentation using perceptual organization and
  location prior. In: CVPR (2015)

\bibitem{NIPS2014_5423}
Goodfellow, I., Pouget-Abadie, J., Mirza, M., Xu, B., Warde-Farley, D., Ozair,
  S., Courville, A., Bengio, Y.: Generative adversarial nets. In: NIPS (2014)

\bibitem{gorelick2007}
Gorelick, L., Blank, M., Shechtman, E., Irani, M., Basri, R.: Actions as
  space-time shapes. IEEE Transactions on Pattern Analysis and Machine
  Intelligence  \textbf{29}(12),  2247--2253 (2007)

\bibitem{HallerL17}
Haller, E., Leordeanu, M.: Unsupervised object segmentation in video by
  efficient selection of highly probable positive features. In: ICCV (2017)

\bibitem{hara2018can}
Hara, K., Kataoka, H., Satoh, Y.: Can spatiotemporal 3d cnns retrace the
  history of 2d cnns and imagenet? In: CVPR (2018)

\bibitem{he2016deep}
He, K., Zhang, X., Ren, S., Sun, J.: Deep residual learning for image
  recognition. In: CVPR (2016)

\bibitem{NIPS2017_7240}
Heusel, M., Ramsauer, H., Unterthiner, T., Nessler, B., Hochreiter, S.: Gans
  trained by a two time-scale update rule converge to a local nash equilibrium.
  In: NIPS (2017)

\bibitem{Huang_2017_CVPR}
Huang, X., Li, Y., Poursaeed, O., Hopcroft, J., Belongie, S.: Stacked
  generative adversarial networks. In: CVPR (2017)

\bibitem{IMSKDB17}
Ilg, E., Mayer, N., Saikia, T., Keuper, M., Dosovitskiy, A., Brox, T.: Flownet
  2.0: Evolution of optical flow estimation with deep networks. In: CVPR (2017)

\bibitem{ioffe2015batch}
Ioffe, S., Szegedy, C.: Batch normalization: Accelerating deep network training
  by reducing internal covariate shift. arXiv preprint arXiv:1502.03167  (2015)

\bibitem{JainXG17}
Jain, S.D., Xiong, B., Grauman, K.: Fusionseg: Learning to combine motion and
  appearance for fully automatic segmentation of generic objects in videos. In:
  CVPR (2017)

\bibitem{Jang2018VideoPW}
Jang, Y., Kim, G., Song, Y.: Video prediction with appearance and motion
  conditions. In: ICML (2018)

\bibitem{pmlr-v80-jang18a}
Jang, Y., Kim, G., Song, Y.: Video prediction with appearance and motion
  conditions. In: ICML (2018)

\bibitem{keuper2015}
Keuper, M., Andres, B., Brox, T.: Motion trajectory segmentation via minimum
  cost multicuts. In: ICCV (2015)

\bibitem{8100267}
Koh, Y.J., Kim, C.: Primary object segmentation in videos based on region
  augmentation and reduction. In: CVPR (2017)

\bibitem{NIPS2017_6639}
Lai, W.S., Huang, J.B., Yang, M.H.: Semi-supervised learning for optical flow
  with generative adversarial networks. In: NIPS (2017)

\bibitem{6126471}
Lee, Y.J., Kim, J., Grauman, K.: Key-segments for video object segmentation.
  In: ICCV (2011)

\bibitem{long2015fully}
Long, J., Shelhamer, E., Darrell, T.: Fully convolutional networks for semantic
  segmentation. In: CVPR (2015)

\bibitem{Mahasseni_2017_CVPR}
Mahasseni, B., Lam, M., Todorovic, S.: Unsupervised video summarization with
  adversarial lstm networks. In: CVPR (2017)

\bibitem{Man+18b}
Maninis, K.K., Caelles, S., Chen, Y., Pont-Tuset, J., Leal-Taix\'e, L.,
  Cremers, D., {Van Gool}, L.: Video object segmentation without temporal
  information. IEEE Transactions on Pattern Analysis and Machine Intelligence
  (2018)

\bibitem{Mao_2017_ICCV}
Mao, X., Li, Q., Xie, H., Lau, R.Y., Wang, Z., Paul~Smolley, S.: Least squares
  generative adversarial networks. In: ICCV (2017)

\bibitem{odena2016semi}
Odena, A.: Semi-supervised learning with generative adversarial networks. arXiv
  preprint arXiv:1606.01583  (2016)

\bibitem{ohnishi2018hierarchical}
Ohnishi, K., Yamamoto, S., Ushiku, Y., Harada, T.: Hierarchical video
  generation from orthogonal information: Optical flow and texture. In: AAAI
  (2018)

\bibitem{OhnishiYUH18}
Ohnishi, K., Yamamoto, S., Ushiku, Y., Harada, T.: Hierarchical video
  generation from orthogonal information: Optical flow and texture. In: AAAI
  (2018)

\bibitem{papazoglou2013fast}
Papazoglou, A., Ferrari, V.: Fast object segmentation in unconstrained video.
  In: ICCV (2013)

\bibitem{8099855}
Perazzi, F., Khoreva, A., Benenson, R., Schiele, B., Sorkine-Hornung, A.:
  Learning video object segmentation from static images. In: CVPR (2017)

\bibitem{7780454}
Perazzi, F., Pont-Tuset, J., McWilliams, B., Gool, L.V., Gross, M.,
  Sorkine-Hornung, A.: A benchmark dataset and evaluation methodology for video
  object segmentation. In: CVPR (2016)

\bibitem{6248065}
{Prest}, A., {Leistner}, C., {Civera}, J., {Schmid}, C., {Ferrari}, V.:
  Learning object class detectors from weakly annotated video. In: CVPR. pp.
  3282--3289 (2012)

\bibitem{RadfordMC15}
Radford, A., Metz, L., Chintala, S.: Unsupervised representation learning with
  deep convolutional generative adversarial networks. ICLR  (2016)

\bibitem{RadosavovicDGGH18}
Radosavovic, I., Doll{\'{a}}r, P., Girshick, R.B., Gkioxari, G., He, K.: Data
  distillation: Towards omni-supervised learning. In: CVPR (2018)

\bibitem{NIPS2017_6797}
Roth, K., Lucchi, A., Nowozin, S., Hofmann, T.: Stabilizing training of
  generative adversarial networks through regularization. In: NIPS (2017)

\bibitem{Saito_2017_ICCV}
Saito, M., Matsumoto, E., Saito, S.: Temporal generative adversarial nets with
  singular value clipping. In: ICCV (2017)

\bibitem{NIPS2016_6125}
Salimans, T., Goodfellow, I., Zaremba, W., Cheung, V., Radford, A., Chen, X.,
  Chen, X.: Improved techniques for training gans. In: Lee, D.D., Sugiyama, M.,
  Luxburg, U.V., Guyon, I., Garnett, R. (eds.) NIPS (2016)

\bibitem{Shoemake:1985}
Shoemake, K.: Animating rotation with quaternion curves. SIGGRAPH  (1985)

\bibitem{ucf101}
Soomro, K., Zamir, A.R., Shah, M.: {UCF101:} {A} dataset of 101 human actions
  classes from videos in the wild  (2012)

\bibitem{Souly_2017_ICCV}
Souly, N., Spampinato, C., Shah, M.: Semi supervised semantic segmentation
  using generative adversarial network. In: ICCV (2017)

\bibitem{BMVC2015_186}
Stretcu, O., Leordeanu, M.: Multiple frames matching for object discovery in
  video. In: BMVC (2015)

\bibitem{43022}
Szegedy, C., Liu, W., Jia, Y., Sermanet, P., Reed, S., Anguelov, D., Erhan, D.,
  Vanhoucke, V., Rabinovich, A.: Going deeper with convolutions. In: CVPR
  (2015)

\bibitem{Tokmakov17}
Tokmakov, P., Alahari, K., Schmid, C.: Learning motion patterns in videos. In:
  CVPR (2017)

\bibitem{Tsai2012}
Tsai, D., Flagg, M., Nakazawa, A., Rehg, J.M.: Motion coherent tracking using
  multi-label mrf optimization. International Journal of Computer Vision
  \textbf{100}(2),  190--202 (2012)

\bibitem{Tulyakov_2018_CVPR}
Tulyakov, S., Liu, M.Y., Yang, X., Kautz, J.: Mocogan: Decomposing motion and
  content for video generation. In: CVPR (2018)

\bibitem{Tzeng_2017_CVPR}
Tzeng, E., Hoffman, J., Saenko, K., Darrell, T.: Adversarial discriminative
  domain adaptation. In: CVPR (2017)

\bibitem{VillegasYHLL17}
Villegas, R., Yang, J., Hong, S., Lin, X., Lee, H.: Decomposing motion and
  content for natural video sequence prediction. ICLR  (2017)

\bibitem{NIPS2016_6194}
Vondrick, C., Pirsiavash, H., Torralba, A.: Generating videos with scene
  dynamics. In: NIPS (2016)

\bibitem{VondrickSFGM18}
Vondrick, C., Shrivastava, A., Fathi, A., Guadarrama, S., Murphy, K.: Tracking
  emerges by colorizing videos. In: ECCV (2018)

\bibitem{Vondrick_2017_CVPR}
Vondrick, C., Torralba, A.: Generating the future with adversarial
  transformers. In: CVPR (2017)

\bibitem{Vondrick017}
Vondrick, C., Torralba, A.: Generating the future with adversarial
  transformers. In: CVPR (2017)

\bibitem{wang2018vid2vid}
Wang, T.C., Liu, M.Y., Zhu, J.Y., Liu, G., Tao, A., Kautz, J., Catanzaro, B.:
  Video-to-video synthesis. In: NeurIPS (2018)

\bibitem{7298961}
Wang, W., Shen, J., Porikli, F.: Saliency-aware geodesic video object
  segmentation. In: CVPR (2015)

\bibitem{xie2017aggregated}
Xie, S., Girshick, R., Doll{\'a}r, P., Tu, Z., He, K.: Aggregated residual
  transformations for deep neural networks. In: CVPR (2017)

\bibitem{Yi_2017_ICCV}
Yi, Z., Zhang, H., Tan, P., Gong, M.: Dualgan: Unsupervised dual learning for
  image-to-image translation. In: ICCV (2017)

\bibitem{Zhang_2017_ICCV}
Zhang, H., Xu, T., Li, H., Zhang, S., Wang, X., Huang, X., Metaxas, D.N.:
  Stackgan: Text to photo-realistic image synthesis with stacked generative
  adversarial networks. In: ICCV (2017)

\bibitem{Zhu_2017_ICCV}
Zhu, J.Y., Park, T., Isola, P., Efros, A.A.: Unpaired image-to-image
  translation using cycle-consistent adversarial networks. In: ICCV (2017)

\end{thebibliography}
\end{document}